\documentclass[11pt]{article}

\usepackage[final]{acl}
\usepackage{times}
\usepackage{latexsym}
\usepackage{fontawesome}

\usepackage[T1]{fontenc}

\usepackage[utf8]{inputenc}

\usepackage{microtype}

\usepackage{inconsolata}

\usepackage{graphicx}

\usepackage{amsfonts}	
\usepackage{xspace}
\usepackage{hyperref}

\usepackage{amsmath}
\usepackage{amssymb}

\usepackage{xcolor}

\usepackage[most]{tcolorbox}
\usepackage{booktabs}
\usepackage{graphicx}   
\usepackage{xcolor}     
\usepackage[most]{tcolorbox} 
\usepackage{pifont}     
\usepackage{graphicx}
\usepackage{tabularx}
\usepackage{booktabs}
\usepackage[table]{xcolor}  

\usepackage{booktabs,tabularx,colortbl,xcolor,array}
\definecolor{HdrBG}{HTML}{EDE7DB}   
\definecolor{SectBG}{HTML}{E3DBC9}  
\definecolor{AltBG}{HTML}{F8F4EC}   
\definecolor{Rule}{HTML}{D5CDBE}    
\definecolor{cure}{HTML}{FFE9B3}    

\arrayrulecolor{Rule}
\newcolumntype{Y}{>{\centering\arraybackslash}X}

\usepackage{stackengine,graphicx}
\newcommand\drsh{\mathop{\ensurestackMath{%
  \stackengine{-1.2pt}{\rightarrow}{\scalebox{1}[.35]{%
    $\mkern-1.3mu\vert$}}{O}{l}{F}{F}{S}}}}

\usepackage{array}
\usepackage{tabularx}
\usepackage{booktabs}
\usepackage[table]{xcolor}
\arrayrulecolor{black}

\tcbset{
  examplebox/.style={
    colback=white,
    colframe=black,
    arc=2mm,
    boxsep=3pt,
    left=2pt,right=2pt,top=2pt,bottom=2pt
  }
}
\usepackage{multirow}
\usepackage{float}
\usepackage{xcolor}
\usepackage{tcolorbox}
\tcbuselibrary{breakable}
\usepackage{fvextra}
\usepackage{balance}
\usepackage{placeins}

\newtcolorbox{promptbox}{
  colback=gray!10,
  colframe=black,
  arc=2pt,
  boxrule=0.5pt,
  left=6pt,
  right=6pt,
  top=6pt,
  bottom=6pt,
  breakable
}

\newcommand{\bigxmark}{\textcolor{red}{\ding{55}}}   
\newcommand{\bigcmark}{\textcolor{green!50!black}{\ding{51}}} 

\newcommand{\method}{\textsc{Cure-Med}\xspace}
\newcommand{\methodBench}{\textsc{CureMed-Bench}\xspace}

\newcommand{\eg}{\textit{e.g., \xspace}}
\newcommand{\xhdr}[1]{\vspace{0em}\noindent{{\bf #1.}}}
\newcommand{\std}[1]{\mathbin{\scriptstyle\pm#1}}

\usepackage[toc,page,header]{appendix}
\usepackage{minitoc}
\doparttoc

\title{\method: Curriculum-Informed Reinforcement Learning for Multilingual Medical Reasoning}

\author{
  \textbf{Eric Onyame}$^{1*}$ \quad
  \textbf{Akash Ghosh}$^{2*}$ \quad
  \textbf{Subhadip Baidya}$^{3}$ \quad
  \textbf{Sriparna Saha}$^{2}$ \\
  \textbf{Xiuying Chen}$^{4}$ \quad
  \textbf{Chirag Agarwal}$^{1}$ \\[0.5em]
  University of Virginia$^{1}$ \quad Indian Institute of Technology Patna$^{2}$ \\  \quad Indian Institute of Technology Kanpur$^{3}$ \quad MBZUAI$^{4}$
}

\begin{document}
\faketableofcontents 

\maketitle

\renewcommand{\thefootnote}{\fnsymbol{footnote}}
\footnotetext[1]{\raggedright Equal contribution}
\footnotetext[2]{\raggedright Corresponding author: \href{mailto:reh6ed@virginia.edu}{reh6ed@virginia.edu}}



\begin{abstract}

While large language models (LLMs) have shown to perform well on monolingual mathematical and commonsense reasoning, they remain unreliable for multilingual medical reasoning applications, hindering their deployment in multilingual healthcare settings.
We address this by first introducing \methodBench, a high-quality multilingual medical reasoning dataset with open-ended reasoning queries with a single verifiable answer, spanning thirteen languages, including underrepresented languages such as Amharic, Yoruba, and Swahili. 
Building on this dataset, we propose \method, a curriculum-informed reinforcement learning framework that integrates code-switching-aware supervised fine-tuning and Group Relative Policy Optimization to jointly improve logical correctness and language stability. Across thirteen languages, our approach consistently outperforms strong baselines and scales effectively, achieving 85.21\% language consistency and 54.35\% logical correctness at 7B parameters, and 94.96\% language consistency and 70.04\% logical correctness at 32B parameters. These results support reliable and equitable multilingual medical reasoning in LLMs.
The code and dataset are available at \href{https://cure-med.github.io/}{\texttt{cure\_med}}.

\end{abstract}

\section{Introduction}
\label{sec:intro}
\looseness=-1 Recent progress in large language models (LLMs) and reasoning-oriented systems has produced strong performance in mathematical reasoning and code generation~\citep{li2022competition, liu2024deepseek, wei2022chain, huang2022towards}. While these advances suggest LLMs can learn structured solution strategies beyond pattern completion, medical reasoning remains challenging~\citep{magrabi2019artificial, stead2018clinical,ghosh2025multilingual,ghosh2026carepilot} because it requires domain knowledge, careful use of context, and reasoning that clinicians can inspect~\citep{patel2005thinking, arocha2005identifying}.

\looseness=-1 Prior work shows promising medical QA and text generation tasks, yet reliable medical reasoning still depends on reasoning-centric data and evaluations that test reasoning behavior rather than answer plausibility~\citep{lievin2024can, singhal2025toward, nori2023capabilities,ghosh2024medsumm,ghosh2024healthalignsumm,ghosh2024clipsyntel}. Without such resources, models may generate fluent, credible-sounding outputs without dependable reasoning. The problem is amplified in multilingual settings: progress remains English-centered, leaving mid- and low-resource languages underrepresented and reliability uneven across communities\cite{ghosal2025relic,ghosh2025multilingual,singh2025let,maji2025drishtikon}. Despite cross-lingual transfer, open-ended medical reasoning often exhibits two recurring failures: \textit{reduced logical accuracy} and \textit{unstable language behavior}~\citep{cahyawijaya2024llms, nguyen2023democratizing}. For clinical use, these failures erode interpretability and trust, since clinicians and patients must understand not only what a system concludes, but how it arrives there~\citep{amann2020explainability}.

\begin{figure*}[t]
    \centering
    \includegraphics[width=\textwidth]{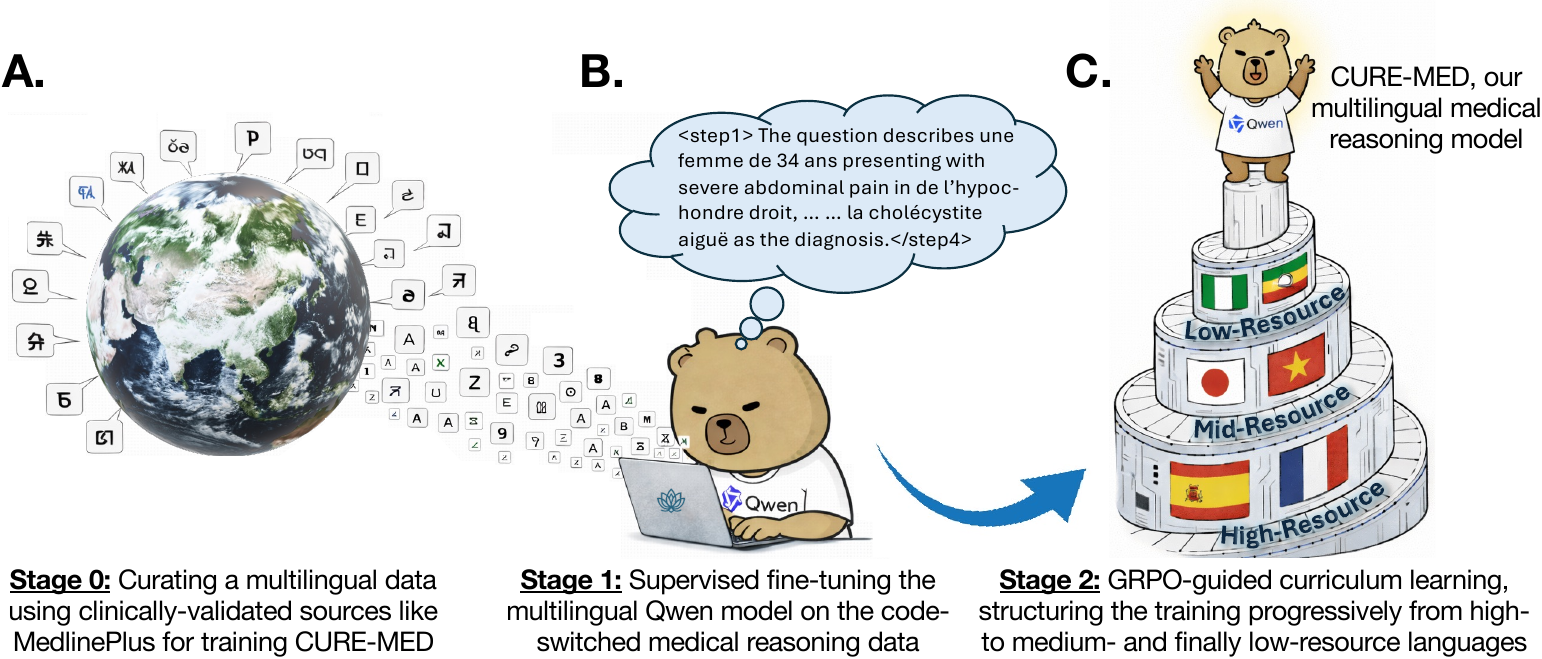}
    \caption{\xhdr{The \method pipeline for multilingual medical reasoning} The framework progresses through three stages: (A) curation of clinically validated multilingual data from sources like MedlinePlus to enable cross-lingual reasoning; (B) supervised fine-tuning of the Qwen2.5-Instruct backbone on code-switched reasoning traces; and (C) GRPO-guided curriculum reinforcement learning, progressively training from high- to mid- and low-resource languages to enhance logical correctness and language consistency.}
    \label{fig:main_figure}
\end{figure*}

While recent efforts attempt to strengthen medical capability through domain-specific supervision~\citep{liu2024survey, shengyu2023instruction}, benchmarks primarily remain monolingual and rely on closed-form settings, providing limited visibility into multilingual reasoning quality and language fidelity~\citep{qiu2024towards}. As LLMs increasingly support clinical education and decision-making, systematic evaluation of multilingual reasoning and language consistency becomes essential for fairness, reliability, and generalization~\citep{lievin2024can, cahyawijaya2024llms}.

In this work, We study multilingual medical reasoning across 13 high-, mid-, and low-resource languages. We introduce \methodBench, an open-ended benchmark where each query has a single verifiable answer, enabling independent evaluation of logical accuracy and language consistency and analysis of cross-lingual generalization under clinically grounded constraints.Next, we propose \textsc{CureMed}, a two-stage training framework (see Figure~\ref {fig:main_figure}) for multilingual medical reasoning. We apply code-switching-aware supervised fine-tuning (SFT) to stabilize language usage during intermediate reasoning steps and perform curriculum-informed GRPO to improve logical correctness and language fidelity. 
Our contributions are: \textbf{1)} We present a systematic evaluation of multilingual medical reasoning of LLMs using verifiable medical queries, enabling reliable measurement of logical accuracy and language consistency across languages; \textbf{2)} We introduce \textsc{CureMed-Bench}, a large-scale multilingual medical reasoning dataset spanning 13 languages across high-, mid-, and low-resource settings; \textbf{3)} We propose \method, a two-stage training framework for multilingual medical reasoning that combines code-switching-aware SFT with curriculum-informed reinforcement learning (RL) to jointly optimize logical correctness and linguistic fidelity; and \textbf{4)} Through extensive automatic and human evaluations, we show that \method achieve state-of-the-art performance on \methodBench and demonstrate improved out-of-distribution generalization, including improved robustness in low-resource languages and stronger performance on unseen medical questions and languages.

\section{Related Work}
\label{sec:related}
\looseness=-1 This work lies at the intersection of medical reasoning with LLMs and multilingual reasoning. We summarize key gaps in prior work and position \method as a unified response.

\looseness=-1\xhdr{Large Medical Reasoning Models} LLMs have been widely studied for medical QA, clinical retrieval, and diagnostic tasks \citep{guo2022medical, singhal2025toward, liu2024survey}. Domain-specific pretraining and instruction tuning can improve factuality, yet benchmark gains often do not translate to reliable medical reasoning \citep{nori2023capabilities, chen2025benchmarking}, with models producing fluent but clinically unsound explanations \citep{amann2020explainability}.
A core issue is evaluation: many medical benchmarks are closed-form (e.g., multiple-choice), which hides intermediate reasoning and limits verification of logical validity \citep{chen2025benchmarking, qiu2024towards}. Recent open-ended evaluations exist, but are largely monolingual or limited to a few high-resource languages, leaving multilingual medical reasoning underexplored \citep{qiu2024towards, schmidgall2024addressing}.

\looseness=-1\textit{We address these gaps by introducing open-ended medical queries with single verifiable answers across 13 diverse languages, enabling independent assessment of reasoning correctness.}\vspace{0.05in}

\looseness=-1\xhdr{Multilingual Reasoning and Language Fidelity} Prior work shows CoT prompting can enable cross-lingual inference transfer \citep{wei2022chain, shi2022language, kojima2022large}, but evaluations mostly target general-domain math/symbolic tasks and skew toward high-resource languages \citep{huang2022towards, chen2023breaking, she2024mapo, nguyen2023democratizing, cahyawijaya2024llms}. In medical settings, models often exhibit degraded accuracy, language drift, and weak cross-lingual generalization \citep{qiu2024towards, schmidgall2024addressing}. Methods such as language mixing and supervised reasoning distillation can improve fluency, but are typically studied in limited bilingual settings or overfit high-resource languages \citep{hammerl2022speaking, yoo2024code, ge2023supervised, huang2024o1, ye2025limo}. RL has also been used to promote structured reasoning, but remains largely English-centric and general-domain \citep{ouyang2022training, achiam2023gpt, jaech2024openai, guo2025deepseek, luong2024reft}.

\textit{\method differs from prior work by optimizing language fidelity and reasoning correctness jointly. We evaluate across high-, mid-, and low-resource languages, and integrate code-switching-aware supervision with curriculum-informed RL for robust multilingual medical reasoning.}

\section{Methodology}
\label{sec:method}

Here, we describe the construction of \methodBench (Sec.~\ref{sec:dataset}), including dataset collection and human verification. Next, we present \method: cold-start initialization (Sec.~\ref{sec:cold}), reward design (Sec.~\ref{sec:reward-design}), and GRPO-guided curriculum reinforcement learning (Sec.~\ref{sec:curriculum}).

\subsection{Dataset Collection}
\label{sec:dataset}
We construct \methodBench, a multilingual medical reasoning dataset of 15{,}774 open-ended QA instances across 13 languages spanning Africa, Asia, and Europe, enabling evaluation under diverse linguistic conditions (including African languages such as Hausa, Yoruba, and Swahili). A breakdown by language and language family is provided in Appendix~\ref{sec:dataset_composition}.

\looseness=-1\xhdr{Source Material and Question Generation} \methodBench is grounded in \textit{MedlinePlus}, a clinically validated medical resource curated by U.S. federal health agencies. Following tool-assisted synthetic data generation~\citep{parisi2022talm,taori2023stanford,zhou2023lima,wang2022super,schick2023toolformer,ghosh2025clinic}, we use GPT-4o to retrieve MedlinePlus content and draft closed-ended multiple-choice questions in each target language. Each item is anchored to the source, includes four options with exactly one correct answer, and provides clinically grounded supervision prior to conversion to open-ended prompts.

\looseness=-1\xhdr{Filtering for Reasoning Difficulty} Following~\citet{chen2024huatuogpt}, we apply multi-stage filtering to retain questions requiring substantive medical reasoning. We remove trivial items by discarding questions answered correctly by all three compact LLMs: Qwen2.5-3B/7B~\citep{xu2025qwen2} and LLaMA-3.1-8B~\citep{grattafiori2024llama}. We further exclude under-specified or ambiguous questions, retaining samples with a single, unambiguous correct answer and consistent cross-lingual interpretation; GPT-4o is used to identify cases with multiple valid answers or cross-lingual inconsistency.

\xhdr{Conversion to Open-Ended Problems} We convert each remaining item into an open-ended prompt \(x\) using GPT-4o, and generate an explicit reasoning chain \(r\) with a free-form ground-truth answer \(y^*\). This removes multiple-choice cues and yields open-ended instances with supervised reasoning, enabling direct evaluation of reasoning quality and answer correctness. We define the dataset as $\mathcal{D}=\{(x,r,y^*)\}$, where each instance has a single clinically grounded solution supported by an explicit reasoning trace. As summarized in Table~\ref{tab:medical-benchmarks}, \methodBench contains 15{,}774 instances across 13 languages, including low-resource languages, extending prior benchmarks that are largely multiple-choice and/or linguistically limited.

\xhdr{Human Verification and Ethical Review} All samples are verified by native speakers and medical experts (physicians, advanced medical students, and nursing PhD candidates). Reviewers assess clinical correctness, linguistic fidelity, and cultural appropriateness, revising culture-specific terminology and removing translation artifacts or medically inappropriate content. {Across 13 languages, user studies report an average rating of \textbf{4.89/5}, supporting clinical validity (Appendix Table~\ref{tab:human_verification_scores})}. All procedures were approved by an Institutional Review Board for social and behavioral sciences and followed established ethical research standards. Additional details are provided in Appendix~\ref{app:dataset_details}.

\begin{table}[t]
\centering
\scriptsize
\setlength{\tabcolsep}{0.9pt}
\definecolor{GroupBlue}{RGB}{225,235,245}

\begin{tabular}{p{2cm}ccccc}
\toprule
\rowcolor{GroupBlue}
{Dataset} &
{Lang.} &
{Size} &
\shortstack{{Open-}\\{ended?}} &
\shortstack{{Reasoning}\\{Supervision}} &
\shortstack{{Low-}\\{resource?}} \\
\midrule
MMedBench & 6 & 8.5k & \bigxmark & \bigcmark & \bigxmark \\
MedQA & 3 & 13k & \bigxmark & \bigxmark & \bigxmark \\
MedExpQA & 4 & 2,488 & \bigxmark & \bigcmark & \bigxmark \\
PubMedQA & 1 & 211k & \bigxmark & \bigcmark & \bigxmark \\
MedQAUSMLE & 1 & 11.4k & \bigxmark & \bigxmark & \bigxmark \\
MedMCQA & 1 & 193k & \bigxmark & \bigcmark & \bigxmark \\
\shortstack[l]{OphthaLingua} & 7 & 1,184 & \bigxmark & \bigxmark & \bigcmark \\
\shortstack[l]{MCMLE} & 1 & 270k & \bigxmark & \bigxmark & \bigxmark \\
XMedBench & 4 & 8,280 & \bigxmark & \bigxmark & \bigxmark \\
WorldMedQA & 4 & 568 & \bigxmark & \bigxmark & \bigxmark \\
HealthSearchQA & 1 & 3,173 & \bigcmark & \bigcmark & \bigxmark \\
\rowcolor{GroupBlue}
\textbf{\shortstack[l]{\method-\\Bench}} & \textbf{13} & \textbf{15,774} & \bigcmark & \bigcmark & \bigcmark \\
\bottomrule
\end{tabular}

\vspace{-2mm}
\caption{Comparison of medical domain benchmarks.}
\label{tab:medical-benchmarks}
\vspace{-2mm}
\end{table}

\subsection{Cold-Start Initialization via Supervised Fine-Tuning (SFT)}
\label{sec:cold}
We initialize multilingual reasoning with a cold-start SFT stage on \emph{code-switched long CoT} trajectories. This stage stabilizes multi-step reasoning in the base model before we introduce stricter language-consistency constraints in later training. Given an input query \(x\) in the target language \(\ell\), we construct a multi-step reasoning trajectory that allows controlled code-switching in intermediate steps (see Figure~\ref {fig:coldstart_example} for a French subset example). Each trajectory contains reasoning steps $\mathbf{r} = \{ r_1, \dots, r_T \}$, where step \(r_t\) may be written in language \(\ell_t \in \mathcal{L}\), followed by a final answer \(y^*\) written in the target language \(\ell\).

We fine-tune the model by maximizing the likelihood of the reasoning trajectory and final answer conditioned on the input: $\mathcal{L}_{\text{SFT}} = -\log p_\theta(\mathbf{r}, y^* \mid x),$
training the model to produce multi-step reasoning before generating the final response. Code-switching in \(\mathbf{r}\) allows the LLM use the most effective language for intermediate inference while keeping the final answer in \(\ell\). The resulting language-adaptive reasoning behavior provides a strong initialization for RL stages that enforce language consistency without degrading logical accuracy.

\begin{figure*}[t]
\centering
\begin{tcolorbox}[
  width=\textwidth,
  colback=gray!5,
  colframe=black,
  boxsep=2pt,
  left=3pt,
  right=3pt,
  top=2pt,
  bottom=2pt,
  title=\textbf{Example: Cold-Start Code-Switched Reasoning (French)}
]
\scriptsize
\textbf{Question.} \textcolor{red}{Une femme de 34 ans se présente aux urgences avec des douleurs abdominales sévères localisées dans l'hypochondre droit, associées à des nausées et des vomissements. Étant donné ses antécédents de calculs biliaires, quel est le diagnostic le plus probable pour son état?}

\vspace{2pt}
\textbf{Reasoning (code-switched CoT).}
\texttt{<thinking>} 
\textcolor{blue}{\texttt{<step1>}} The question describes \textcolor{red}{une femme de 34 ans} presenting with severe abdominal pain in the \textcolor{red}{région de l'hypochondre droit}, along with \textcolor{red}{des nausées et des vomissements}. These are classic upper abdominal symptoms. \textcolor{blue}{\texttt{</step1>}}
\textcolor{blue}{\texttt{<step2>}} \textcolor{red}{Ses symptômes suggèrent une atteinte possible de la vésicule biliaire}, especially given her history of \textcolor{red}{calculs biliaires}. This could indicate a gallbladder-related condition. \textcolor{blue}{\texttt{</step2>}}
\textcolor{blue}{\texttt{<step3>}} The presence of \textcolor{red}{fièvre à 38,5°C} and tenderness upon palpation of the \textcolor{red}{quadrant supérieur droit} indicates an inflammatory process --- likely \textcolor{red}{une infection}. \textcolor{blue}{\texttt{</step3>}}
\textcolor{blue}{\texttt{<step4>}} In this context, \textcolor{red}{on peut fortement envisager la cholécystite aiguë} (acute cholecystitis) as the most probable diagnosis given the presentation and risk factors. \textcolor{blue}{\texttt{</step4>}}
\texttt{</thinking>}

\vspace{2pt}
\textbf{Answer.}
\texttt{<answer>}
\textcolor{green!50!black}{Le diagnostic le plus probable est la cholécystite aiguë}
\texttt{</answer>}

\vspace{1pt}
\hrule
\vspace{1pt}
\textbf{Language:} \textcolor{red}{French} \hfill \textbf{Type:} Cold-start code-switched CoT sample
\end{tcolorbox}
\vspace{-2mm}
\caption{\looseness=-1 An example from the cold-start multilingual dataset showing CoT reasoning in French. The reasoning combines English-based clinical terms and local-language expressions, reflecting code-switching in medical contexts.}
\label{fig:coldstart_example}
\end{figure*}


\subsection{Reward Design}
\label{sec:reward-design}
We train \method with a weighted reward that promotes clinical correctness, language fidelity, and adherence to a structured output format. We use a closed-source multilingual reward model that performs competitively on RewardBench \citep{lambert2025rewardbench}. To mitigate same-model judge bias, we use a separate model for LLM-as-a-judge verification \citep{verga2024replacing, bansal2023peering}.\vspace{0.03in}

\looseness=-1\xhdr{Correctness Reward} Following~\citet{zheng2023judging}, we use GPT-4.1 as a verifier to score semantic and clinical equivalence between the model output ($y$), and reference answer ($y^{\ast}$). The verifier returns a continuous score in $[0,1]$:
\begin{equation}
R_{\text{acc}}(y \mid x, y^{\ast}) = v_{\text{acc}}(x, y, y^{\ast}) \in [0,1].
\end{equation}
We use exact-match scoring for closed-ended questions. For open-ended questions, the verifier assigns partial credit when the response reaches the correct conclusion via clinically valid reasoning, even under paraphrase~\citep{su2025crossing}, providing smoother learning signals.

\xhdr{Language Consistency Reward} We enforce strict output-language fidelity by scoring whether $y$ is written entirely in the query language $\ell$:
\begin{equation}
R_{\text{lang}}(y \mid \ell) =
\begin{cases}
1 & \text{if the language of } y \text{ matches } \ell \\
0 & \text{otherwise.}
\end{cases}
\label{eq:lang_reward}
\end{equation}

\xhdr{Format Reward} A parser checks compliance with the required structure (\texttt{<thinking>}, numbered \texttt{<step n>}, and \texttt{<answer>} tags):
\begin{equation}
R_{\text{fmt}}(y) =
\begin{cases}
1 & \text{if the required format is followed} \\
0 & \text{otherwise.}
\end{cases}
\label{eq:fmt_reward}
\end{equation}

\noindent The final composite reward is defined as:
\begin{equation}
\begin{split}
R(y \mid x, y^{\ast}, \ell) = & \; \lambda_{\text{acc}} R_{\text{acc}}(y \mid x, y^{\ast}) + \\& \lambda_{\text{lang}} R_{\text{lang}}(y \mid \ell) + \lambda_{\text{fmt}} R_{\text{fmt}}(y)
\end{split}
\label{eq:reward_combined}
\end{equation}

\begin{figure*}[t]
  \centering
  \begin{tcolorbox}[
    examplebox,
    width=\textwidth,
    boxsep=2pt,
    left=3pt,
    right=3pt,
    top=2pt,
    bottom=2pt,
    title=\textbf{Example: Baseline vs.\ CURE-Med (Spanish)}
  ]
  \scriptsize
  \textbf{Question.}
  \textit{Un paciente presenta congestión nasal y tos leve desde hace dos días.
  No tiene fiebre ni dificultad para respirar. ¿Cuál es la causa más probable?}
  
  \vspace{3pt}
  \hrule
  \vspace{3pt}

  \begin{minipage}[t]{0.48\linewidth}
    \raggedright
    \textbf{Baseline model (incorrect)}\\[1pt]
    \textbf{Reasoning (flawed).}
    El cuadro parece un resfriado común, pero
    \textcolor{red}{la ausencia de fiebre podría indicar que no es viral y la tos
    podría ser señal de algo más serio como una infección pulmonar temprana. La
    congestión nasal podría ser un síntoma inicial de una patología más grave.}\\[2pt]
    \textbf{Answer.}
    \textcolor{red}{Podría tratarse de una infección pulmonar temprana.} \bigxmark
  \end{minipage}
  \hfill
  \begin{minipage}[t]{0.48\linewidth}
    \raggedright
    \textbf{CURE-Med (correct)}\\[1pt]
    \textbf{Reasoning (code-switched CoT).}
    \texttt{<step1>} The symptoms are mild,
    \textcolor{blue}{\texttt{lo que coincide con un resfriado leve.}}
    \texttt{</step1>}
    \texttt{<step2>} No fever,
    \textcolor{blue}{\texttt{lo que reduce la probabilidad de neumonía.}}
    \texttt{</step2>}
    \texttt{<step3>}
    \textcolor{blue}{\texttt{Lo más probable es un resfriado viral leve.}}
    \texttt{</step3>}\\[2pt]
    \textbf{Answer.}
    \textcolor{green!50!black}{\texttt{Lo más probable es un resfriado viral leve.}} \bigcmark
  \end{minipage}
  \end{tcolorbox}
  \vspace{-2mm}
  \caption{\looseness=-1 Qualitative Spanish medical-reasoning example comparing a baseline Qwen2.5-7B-Instruct model and \method-7B. The baseline model produces fluent but clinically flawed reasoning (red) and an incorrect diagnosis, whereas \method generates a structured, code-switched CoT (blue) and arrives at the correct diagnosis (green).}
  \label{fig:curemed-spanish-sidebyside-newq}
\end{figure*}

\subsection{ GRPO-guided curriculum reinforcement learning}
\label{sec:curriculum}
After SFT, we fine-tune the model with curriculum-guided GRPO~\citep{shao2024deepseekmath, guo2025deepseek} for optimizing the reasoning policy under the multilingual verifier-driven reward described in Sec.~\ref{sec:reward-design}.

\xhdr{Curriculum Design} We design the curriculum around language resource availability rather than problem complexity. This is motivated by the observation that models achieve higher reasoning accuracy in high-resource languages, providing more stable reward signals early in reinforcement learning. We therefore treat languages as tasks of increasing difficulty and progress from high$\to$medium$\to$low-resource tiers. Based on baseline performance, we define three tiers: high- (French, Japanese, Spanish, Vietnamese), medium- (Korean, Thai, Turkish, Bengali), and low-resource (Amharic, Yoruba, Hausa, Hindi, Swahili). We start GRPO on the high-resource and progressively expand training to lower-resource tiers. To reduce catastrophic forgetting, we retain a fixed fraction of samples from the previous phase when introducing a new tier. Formally, curriculum phase $C_i$ draws samples from languages in tier $L_i \in \{\text{high}, \text{medium}, \text{low}\}$.



\xhdr{Training Procedure} While following prior works ~\citep{shao2024deepseekmath, guo2025deepseek,hwang2025learn}, we apply GRPO without modifying the optimization rule, the training was designed in curriculum phases. When reward improvements plateau within a tier, we expand sampling to include the next tier while mixing in data from the previous phase to preserve earlier capabilities. At phase $i$, we sample batches from: 
$\mathcal{D}_i = \alpha\, \mathcal{D}_{i-1} + (1-\alpha)\, \mathcal{D}_{L_i},$ where $\mathcal{D}_{L_i}$ denotes data from tier $L_i$, $\mathcal{D}_{i-1}$ is the retained data from phase $i-1$, and $\alpha{=}0.85$ controls the retention ratio. This retention-aware curriculum supports incremental transfer to low-resource languages while maintaining performance.
\newcolumntype{Y}{>{\centering\arraybackslash}X}

\begin{table}[t]
\centering
\small
\setlength{\tabcolsep}{1pt}
\renewcommand{\arraystretch}{0.9}

\definecolor{CUREgreen}{RGB}{226,239,218}
\definecolor{CUREtext}{RGB}{0,110,90}
\definecolor{GroupBlue}{RGB}{225,235,245}

\begin{tabularx}{0.48\textwidth}{lcc}
\toprule
\textbf{Model} &
\textbf{Consistency ($\uparrow$)} &
\textbf{Accuracy ($\uparrow$)} \\
\midrule

\rowcolor{GroupBlue}
\multicolumn{3}{l}{\textbf{Small Models ($\leq$ 3B)}} \\
\cmidrule(lr){1-3}

LLaMA-3.2-3B            & $23.69 \std{0.36}$ & $10.41 \std{0.38}$ \\
Qwen2.5-Instruct-1.5B   & $3.84 \std{0.25}$  & $6.20 \std{0.24}$  \\
Qwen2.5-Instruct-3B     & $8.39 \std{0.42}$  & $10.83 \std{0.60}$ \\

\rowcolor{CUREgreen}
\textbf{\method-Qwen2.5-1.5B} &
$\underline{57.60\std{0.65}}$ &
$\underline{28.32\std{0.35}}$ \\

\rowcolor{CUREgreen}
\textbf{\method-Qwen2.5-3B } &
$\textbf{74.28}\std{0.60}$ &
$\textbf{42.93}\std{0.60}$ \\

\midrule
\rowcolor{GroupBlue}
\multicolumn{3}{l}{\textbf{Medium Models (7--9B)}} \\
\cmidrule(lr){1-3}

BioMistral-7B           & $7.10 \std{0.90}$  & $4.80 \std{0.95}$ \\
Gemma-7B                & $0.37 \std{0.25}$  & $1.23 \std{0.80}$ \\
MedAlpaca-7B            & $3.50 \std{0.90}$  & $2.47 \std{0.95}$ \\
Meditron-7B             & $0.43 \std{0.40}$  & $2.50 \std{1.10}$ \\
Mistral-7B              & $18.70 \std{1.30}$ & $15.23 \std{1.20}$ \\
Apollo2-7B              & $25.63 \std{1.35}$ & $15.93 \std{1.35}$ \\
Qwen2.5-Instruct-7B     & $25.44 \std{0.36}$ & $29.56 \std{0.42}$ \\
LLaMA-3.1-Instruct-8B   & $36.56 \std{0.31}$ & $18.91 \std{0.18}$ \\
HuatuoGPT-o1-8B         & $\underline{67.30 \std{0.14}}$ & $\underline{46.86 \std{0.09}}$ \\
OpenBioLLM-Llama3-8B    & $1.47 \std{0.45}$  & $36.62 \std{0.72}$ \\
MMed-Llama-3-8B         & $21.38 \std{0.56}$ & $28.09 \std{0.62}$ \\
UltraMedical LLaMA-3-8B & $47.03 \std{1.03}$ & $35.29 \std{1.10}$ \\
Ministral-8B            & $46.93 \std{0.45}$ & $42.87 \std{0.21}$ \\
LLaMA-3-8B              & $31.58 \std{0.12}$ & $28.93 \std{0.42}$ \\
Gemma-9B                & $23.22 \std{1.14}$ & $36.97 \std{1.03}$ \\

\rowcolor{CUREgreen}
\textbf{\method-Qwen2.5-7B } &
$\textbf{85.21}\std{0.63}$ &
$\textbf{54.35}\std{0.50}$ \\

\midrule
\rowcolor{GroupBlue}
\multicolumn{3}{l}{\textbf{Large Models ($\geq$ 14B)}} \\
\cmidrule(lr){1-3}

MedAlpaca-13B           & $0.10 \std{0.17}$  & $0.07 \std{0.12}$ \\
Qwen2.5-Instruct-14B    & $35.57 \std{0.38}$ & $41.79 \std{0.39}$ \\
Qwen2.5-Instruct-32B    & $41.51 \std{0.38}$ & $49.69 \std{0.40}$ \\
Qwen2.5-Instruct-72B    & $70.73 \std{1.10}$ & $58.80 \std{1.20}$ \\
LLaMA-3.1-70B           & $75.68 \std{1.01}$ & $54.65 \std{0.31}$ \\
LLaMA-3.3-Instruct-70B  & $79.66 \std{0.32}$ & $60.80 \std{0.72}$ \\
HuatuoGPT-o1-70B        & $\underline{86.79 \std{0.44}}$ & $\underline{66.67 \std{0.24}}$ \\
OpenBioLLM-Llama3-70B   & $70.30 \std{0.43}$ & $51.22 \std{0.41}$ \\
Meditron-70B            & $0.21 \std{0.55}$  & $4.54 \std{0.59}$ \\
MMed-LLaMA-3.1-70B      & $26.49 \std{0.36}$ & $37.85 \std{0.76}$ \\

\rowcolor{CUREgreen}
\textbf{\method-Qwen2.5-14B } &
$90.27 \std{0.31}$ &
$63.74 \std{0.43}$ \\

\rowcolor{CUREgreen}
\textbf{\method-Qwen2.5-32B } &
$\textbf{94.96}\std{0.40}$ &
$\textbf{70.04}\std{0.04}$ \\

\bottomrule
\end{tabularx}

\caption{\looseness=-1 Mean results across 13 languages on 28 baseline models and \method. We observe that \method models outperform models in each parameter scale.
\textbf{Consistency} denotes language consistency and \textbf{Accuracy} denotes logical accuracy.
Best overall results are \textbf{bold}, best baselines are \underline{underlined}.
}
\label{tab:multilingual-medical-main}
\end{table}

\section{Experiments}
\label{sec:exp_sec}
\looseness=-1 Next, we outline the experimental setup, baseline models, training and evaluation procedures used to address key research questions:
\textbf{RQ1)} Does \method improve multilingual medical reasoning over instruction-tuned baselines and their vanilla variants?
\textbf{RQ2)} What is the performance trade-off between language fidelity and medical reasoning accuracy?
\textbf{RQ3)} How does curriculum-guided learning affect performance across model scales?
\textbf{RQ4)} Does \method generalize to unseen medical questions and languages under out-of-distribution evaluation?

\subsection{Experimental Setup}
\label{exp:data_splits}
\paragraph{Dataset and Splits.}
All experiments are conducted on \textsc{CureMed-Bench}, where the dataset is partitioned into 80\% train and 20\% held-out test set. The train set is further divided into 80\% for supervised fine-tuning and 20\% for reinforcement fine-tuning. Dataset construction and filtering procedures are described in Sec.~\ref{sec:method}.


\looseness=-1\xhdr{Baselines} We benchmark \method against 28 baseline models comprising i) general-purpose, including Qwen 2.5-Instruct \citep{qwen2.5}, LLaMA \citep{dubey2024llama}, Gemma \citep{team2024gemma}, Mistral \citep{jiang2023mistral7b}, Apollo2 \citep{zheng2024efficientlydemocratizingmedicalllms}, and Ministral \citep{mistral2024ministral}; and ii) medical-specific, including MedAlpaca \citep{han2025medalpacaopensourcecollection}, Meditron \citep{chen2023meditron}, UltraMedical \citep{zhang2024ultramedical}, HuatuoGPT \citep{zhang2305huatuogpt}, OpenBioLLM \citep{saama2024openbiollm}, BioMistral \citep{labrak2024biomistralcollectionopensourcepretrained}, and MMed-LLaMA \citep{qiu2024towards}.
All models are evaluated in a zero-shot setting across three independent runs.

\xhdr{Model Training and Evaluation} We use Qwen-2.5-\{1.5B,3B,7B,14B,32B\} instruction-tuned models as backbones. Training is performed on eight NVIDIA A100 GPUs in two stages: i) SFT on the multi-step cold-switched dataset for three epochs and ii) language-resource-aware curriculum fine-tuning with GRPO. Reinforcement progresses from high- to low-resource languages, retaining 85\% of data from earlier stages to mitigate catastrophic forgetting. See Appendix~\ref{sec:curriculum_language_tiers} for additional details on our high-/low-resource language definition and the criteria used to assign languages to each group.

\looseness=-1 Following~\citet{chen2024huatuogpt}, we evaluate on the held-out test set using an LLM-as-a-judge framework, with GPT-4o used to match each model output to the known ground-truth answer. We assess \emph{logical accuracy} (LA), defined as the clinical accuracy of the final answer, and \emph{language consistency} (LC), defined as whether the final answer is produced in the question's corresponding target language.
Figure~\ref{fig:curemed-spanish-sidebyside-newq} provides a representative Spanish example, illustrating how curriculum-guided reinforcement improves accuracy while maintaining language consistency compared to a fluent but incorrect baseline. See Appendix~\ref{sec:training_verification} for Additional implementation details. 
\section{Results}
\label{sec:results_discussion}
\looseness=-1 Here, we report results that answer RQ1--RQ4 from Sec.~\ref{sec:exp_sec}. We compare \method to instruction-tuned baselines and analyze language-reasoning trade-offs, scaling under curriculum-guided reinforcement, and out-of-distribution generalization. 


\looseness=-1\xhdr{RQ1) \method outperforms baselines} Table~\ref{tab:multilingual-medical-main} compares \method to three baseline families: general-purpose instruction-tuned LLMs, medical-domain instruction-tuned models, and medical-specialized LLMs. Across scales, \method improves both logical accuracy and target-language consistency. At $\leq$3B, baselines show low correctness and frequent language violations, while \method reaches 42.93\% logical correctness and 74.28\% consistency (3B). At 7--9B, \method improves over the best baseline in logical correctness (54.35\% vs.\ 46.86\%) while maintaining 85.21\% consistency. At $\geq$14B, \method remains best, reaching 70.04\% logical correctness and 94.96\% consistency. Notably, our 32B model is competitive with closed-source systems and outperforms several proprietary models on \methodBench (See Appendix~\ref{sec:closed_source_summary}, \ref{sec:closed_source_model}; Tables~\ref{tab:closed-source-curemed-concise}, \ref{tab:closed-source-logical-by-language}, \ref{tab:closed-source-lang-consistency-by-language}).

\begin{figure*}
\centering
\small
\includegraphics[width=\textwidth]{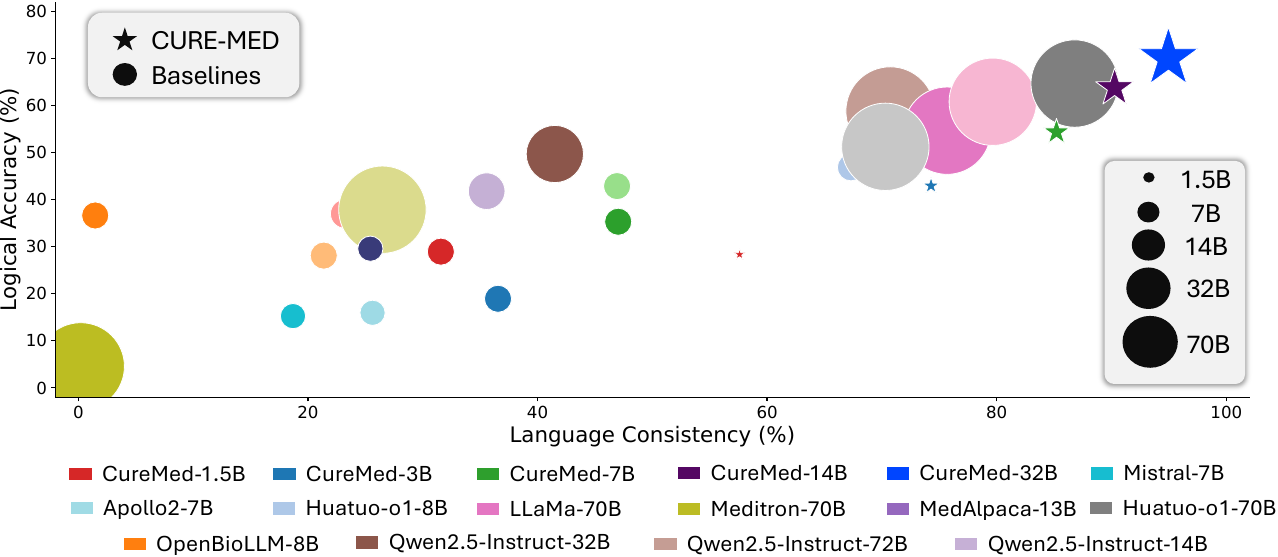}
\caption{\looseness=-1 Trade-off performance between language consistency and logical accuracy of multilingual medical reasoning models, where
each point represents a model instance with bubble size reflecting model scale.
Baseline and \method models are shown as \faCircle~and $\bigstar$, respectively. \method shifts performance toward the upper-right, indicating consistent gains in language consistency and logical accuracy.
}
\label{fig:scatter_plot}
\end{figure*}

\looseness=-1\xhdr{RQ2) \method achieves better language and reasoning trade-offs} Figure.~\ref{fig:scatter_plot} shows that 
while baselines exhibit a weak trade-off between language consistency and logical correctness, 
\method shifts this 
in the upper-right corner, highlighting that \method improves medical reasoning without sacrificing target-language fidelity, addressing a key failure mode of prior multilingual medical systems. We observe that \method-1.5B outperform several baselines ranging from 7B to 70B and our \method-32B model outperform all 28 baseline models.

\begin{figure}[H]
    \centering
\includegraphics[width=0.75\columnwidth]{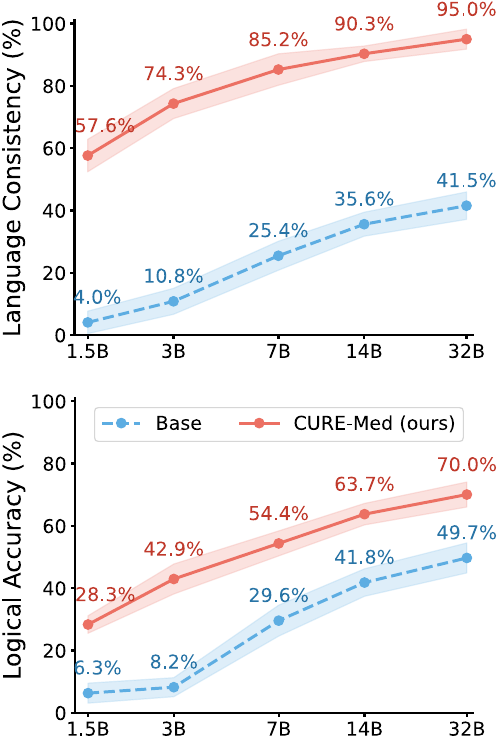}
    \caption{\looseness=-1 Scaling performance of \method vs. base across Qwen2.5-Instruct variants on language consistency \textbf{(top)} and logical accuracy \textbf{(bottom)}. Our method (solid red line) consistently outperforms the base model (dashed blue line), with performance gaps widening at larger model scales, highlighting the effectiveness of \method for multilingual medical reasoning.
    }
    \label{fig:scaling_plots}
\end{figure}

\looseness=-1\xhdr{RQ3) Scaling Trends of \method} Fig.~\ref{fig:scaling_plots} shows that \method smoothly scale language consistency (57.6\%\begin{small}@1.5B\end{small} $\to$ 95.0\%\begin{small}@32B\end{small}) and logical correctness (28.3\%$\to$70.0\%). By comparison, instruction-tuned baselines exhibit only modest gains in language consistency as scale increases, remaining unreliable even at larger scale. Tables~\ref{tab:per_language_7b_base_vs_curemed}-\ref{tab:per_language_3b_base_vs_curemed} in App.~\ref{sec:per_language} report per-language results, showing that \method consistently improves performance across languages and scales effectively. These trends indicate that curriculum-guided reinforcement fundamentally alters scaling behavior by coupling reasoning optimization with language fidelity.

\looseness=-1\xhdr{RQ4) Out-of-distribution cross-lingual generalization} We evaluate transfer to held-out medical benchmarks: 
MMedBench \citep{qiu2024towards}, MedExpQA \citep{alonso2024medexpqa}, and MedQA \citep{jin2021disease}. Across all three benchmarks, \method improves accuracy over the Qwen2.5 backbones in the majority of language--scale settings, with the clearest gains for smaller models. On MMedBench (Table \ref{tab:crosslingual-accuracy}), the 1.5B backbone increases from 6.00$\to$24.00 and from 20.00$\to$57.50 on representative languages, demonstrating strong transfer under limited capacity. MedExpQA (Table \ref{tab:crosslingual-exp-accuracy}) shows a similar large jump at 1.5B, rising from 1.40$\to$44.80, while MedQA (Table \ref{tab:crosslingualqa-accuracy-zh}) improves from 21.00$\to$59.50 at 1.5B on Chinese variants. These gains remain at larger scales, indicating that curriculum-guided RL transfers beyond in-domain training to unseen questions and language variants.

\section{Ablation Study}
Here, we ablate \method's key components and measure their impact on logical accuracy. We also assess robustness by evaluating \method across multiple multilingual medical QA benchmarks and strong medical-domain LLM baselines.

\begin{table*}[h]
\centering
\small
\setlength{\tabcolsep}{2pt}
\renewcommand{\arraystretch}{1.2}

\definecolor{CUREgreen}{RGB}{226,239,218}
\definecolor{GroupBlue}{RGB}{225,235,245}

\begin{tabular}{l c c >{\columncolor{CUREgreen}}c c >{\columncolor{CUREgreen}}c}
\toprule
\textbf{Model size} &
\textbf{Base} &
Na\"ive SFT &
\textbf{\method} (w/o RL) &
Na\"ive RFT &
\textbf{\method} (w/ RL) \\
\midrule

\rowcolor{GroupBlue}
\multicolumn{6}{c}{\textbf{Qwen2.5-Instruct — Language consistency ($\uparrow$)}} \\
\midrule

1.5B & $3.84\std{0.25}$ & $8.60\std{1.23}$ & $53.67\std{0.38}$ (\textbf{+45.07}) & $8.81\std{0.34}$ & ${57.60}\std{0.65}$ (\textbf{+48.79}) \\
3B   & $8.39\std{0.42}$ & $13.07\std{0.33}$ & $72.68\std{0.38}$ (\textbf{+59.61}) & $13.28\std{0.57}$ & ${74.28}\std{0.60}$ (\textbf{+61.00}) \\
7B   & $25.44\std{0.36}$ & $37.11\std{0.44}$ & $83.46\std{0.36}$ (\textbf{+46.35}) & $38.99\std{0.68}$ & ${85.21}\std{0.63}$ (\textbf{+46.22}) \\
14B  & $35.57\std{0.38}$ & $37.20\std{0.33}$ & $84.28\std{0.35}$ (\textbf{+47.08}) & $39.10\std{1.05}$ & ${90.27}\std{0.31}$ (\textbf{+51.17}) \\
32B  & $35.57\std{0.38}$ & $43.00\std{0.27}$ & $90.29\std{0.21}$ (\textbf{+47.29}) & $45.10\std{1.12}$ & ${94.96}\std{0.40}$ (\textbf{+49.86}) \\

\midrule

\rowcolor{GroupBlue}
\multicolumn{6}{c}{\textbf{Qwen2.5-Instruct — Logic accuracy ($\uparrow$)}} \\
\midrule

1.5B & $6.20\std{0.24}$ & $4.61\std{0.36}$ & $22.97\std{0.57}$ (\textbf{+18.36}) & $8.80\std{0.47}$ & ${28.32}\std{0.35}$ (\textbf{+19.52}) \\
3B   & $10.83\std{0.60}$ & $9.50\std{0.38}$ & $39.13\std{0.53}$ (\textbf{+29.63}) & $10.06\std{0.45}$ & ${42.93}\std{0.60}$ (\textbf{+32.87}) \\
7B   & $29.56\std{0.42}$ & $30.05\std{1.10}$ & $50.03\std{0.48}$ (\textbf{+19.98}) & $38.50\std{0.38}$ & ${54.35}\std{0.50}$ (\textbf{+15.85}) \\
14B  & $41.79\std{0.39}$ & $43.10\std{0.13}$ & $61.91\std{0.45}$ (\textbf{+18.81}) & $45.20\std{0.55}$ & ${63.74}\std{0.43}$ (\textbf{+18.54}) \\
32B  & $49.69\std{0.40}$ & $51.21\std{0.15}$ & ${66.34}\std{0.43}$ (\textbf{+15.13}) & $53.40\std{0.49}$ & $70.04\std{0.04}$ (\textbf{+16.64}) \\
\bottomrule
\end{tabular}
\vspace{-0.05in}
\caption{
\looseness=-1 Ablation study of \method.
Results are averaged over three runs and reported as mean $\pm$ standard deviation and green columns denote \method variants. 
}\vspace{-0.1in}
\label{tab:cure_ablation}
\end{table*}

\begin{table}[!htbp]
\centering
\small
\setlength{\tabcolsep}{3pt}
\renewcommand{\arraystretch}{1.1}
\definecolor{CUREgreen}{RGB}{226,239,218}
\definecolor{GroupBlue}{RGB}{225,235,245}
\begin{tabularx}{\linewidth}{l c *{3}{>{\centering\arraybackslash}X}}
\rowcolor{GroupBlue}
\textbf{Model} & \textbf{French} & \textbf{Japanese} & \textbf{Russian} & \textbf{Spanish} \\
Qwen2.5-1.5B         & 6.00  & 11.06 & 20.00 & 20.00 \\
\rowcolor{CUREgreen}
~$\drsh$ \method     & \textbf{24.00} & \textbf{35.18} & \textbf{57.50} & \textbf{44.50} \\
Qwen2.5-3B           & 6.50  & 24.62 & 22.50 & 23.00 \\
\rowcolor{CUREgreen}
~$\drsh$ \method     & \textbf{42.00} & \textbf{37.69} & \textbf{60.50} & \textbf{56.00} \\
Qwen2.5-7B           & 42.00 & \textbf{51.76} & 53.50 & 63.00 \\
\rowcolor{CUREgreen}
~$\drsh$ \method     & \textbf{50.00} & 46.73 & \textbf{66.00} & \textbf{64.00} \\
Qwen2.5-14B          & 61.00 & 57.29 & 63.00 & 71.50 \\
\rowcolor{CUREgreen}
~$\drsh$ \method     & \textbf{64.00} & \textbf{65.83} & \textbf{75.50} & \textbf{78.00} \\
Qwen2.5-32B          & 69.50 & 67.84 & 72.00 & 29.50 \\
\rowcolor{CUREgreen}
~$\drsh$ \method     & \textbf{78.50} & \textbf{77.29} & \textbf{80.00} & \textbf{82.50} \\
\hline
\end{tabularx}
\vspace{-0.05in}
\caption{OOD accuracy on MMedBench. \method improves reasoning performance across all model sizes, showing strong cross-lingual generalization to unseen medical questions and languages. See  Tables~\ref{tab:crosslingual-exp-accuracy}-\ref{tab:crosslingualqa-accuracy-zh} for results on MedExpQA and MedQA datasets.
}\vspace{-0.1in}
\label{tab:crosslingual-accuracy}
\end{table}

\xhdr{Effect of Codeswitched Supervised Fine-Tuning} We isolate the effect of code-switched supervision during SFT by contrasting the base model, naïve SFT trained on multilingual long-CoT data, and \method SFT without reinforcement learning. Naïve SFT yields small and sometimes unstable improvements: language consistency rises from 8.39\%$\to$13.07\% at 3B, yet logic accuracy decreases from 10.83\%$\to$9.50\%, indicating that multilingual instruction tuning does not consistently strengthen medical reasoning as shown in Table~\ref{tab:cure_ablation}. In contrast, code-switched SFT in \method produces large, consistent gains across model scales. At 1.5B, language consistency increases from 3.84\%$\to$53.67\% and logic accuracy from 6.20\%$\to$22.97\%. 
These improvements persist as scale increases, reaching 90.29\% language consistency and 66.34\% logic accuracy at 32B. In summary, the results show that structured code-switching during SFT drives the strongest gains, while naïve multilingual SFT remains insufficient for reliable multilingual medical reasoning.

\looseness=-1\xhdr{Effect of GRPO-guided curriculum reinforcement learning} We assess whether RL adds value beyond SFT by comparing naïve single stage GRPO based RFT against the curriculum and language resource-aware RL used in \method, with results summarized in Table~\ref{tab:cure_ablation}. Naïve RFT yields limited and uneven gains, especially at smaller scales, suggesting that uniform reinforcement signals do not consistently shape multilingual behavior. 
In contrast, \method applies RL after code switched SFT and delivers reliable improvements in both language consistency and logical accuracy across all model sizes. These results show that curriculum and resource-aware RL stabilizes optimization and strengthens multilingual medical reasoning beyond naïve GRPO.

\begin{figure}[t]
    \centering
\includegraphics[width=\columnwidth]{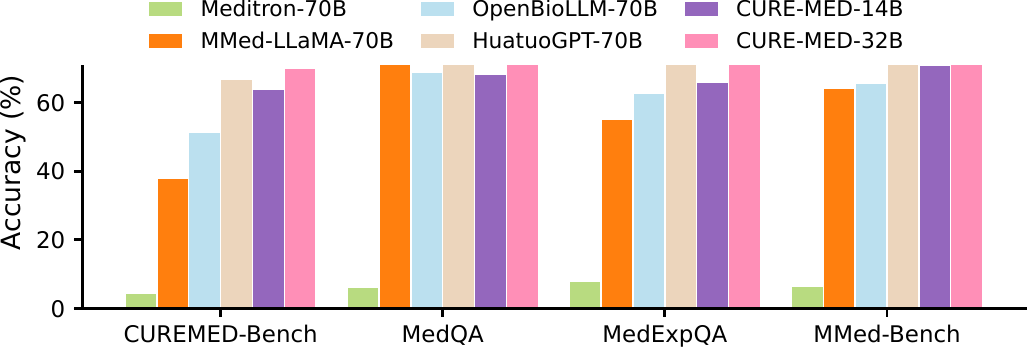}
    \caption{\method vs. medical LLM baselines across four multilingual medical QA benchmarks. Results show logi cal accuracy, highlighting \method's consistent across diverse evaluation settings.
    }\vspace{-0.05in}
    \label{fig:medical_baselines}
\end{figure}

\xhdr{\method vs. Medical LLM baselines across Benchmarks} We evaluate \method against strong medical-domain LLM baselines across four multilingual medical benchmarks (see Fig.~\ref{fig:medical_baselines}).
\method remains consistent, with \method-32B achieving the best performance on \methodBench (70.04\%) and MMed-Bench (79.57\%), and remains competitive on MedQA and MedExpQA, where HuatuoGPT-70B leads narrowly. \method-14B also provides strong results across all benchmarks, while other medical baselines lag behind more substantially, highlighting \method's robustness across diverse evaluation settings.





\section{Conclusion}
\label{sec:conclusion}
We introduce \methodBench, a multilingual medical reasoning benchmark of open-ended questions with explicit reasoning traces and a single verifiable answer across 13 languages, including low-resource settings. Using \methodBench, we propose \method, which combines cold-start code-switched initialization, structured supervised fine-tuning, and language-resource-aware curriculum-RL to improve reasoning while preserving target-language fidelity. Across languages, datasets, and model scales, \method improves logical correctness and language consistency over strong baselines; ablations show supervised and RL stages provide complementary gains for stable multilingual reasoning.

\section{Limitations} \methodBench is constrained by the availability of clinically reliable source material across languages, which limits coverage and can create uneven difficulty between high- and low-resource settings. Our benchmark targets open-ended questions with a single verifiable answer and thus does not capture longitudinal care trajectories, multi-visit decision-making, or multimodal clinical evidence. In addition, parts of our pipeline rely on API-based models (e.g., for generation and/or verification), which can be costly and may hinder reproducibility for some researchers; a practical direction is to replace these components with smaller open-source models trained for the same roles and to release prompts, code, and verifier alternatives to reduce dependence on paid APIs. Future work will expand language coverage, broaden clinical settings and modalities, and further reduce reliance on proprietary APIs.


\section{Ethical Considerations}
\label{sec:ethics}

\looseness=-1 This work supports the evaluation and training of multilingual medical reasoning systems by measuring reasoning correctness and target-language fidelity across diverse languages. \methodBench is derived from publicly available, clinically curated sources and contains no patient records or personally identifiable information. Native speakers and medical experts reviewed all samples for clinical correctness, linguistic fidelity, and cultural appropriateness under IRB-approved procedures, and we report per-language results to surface reliability differences across resource levels.

\section{Acknowledgment}
We would like to thank all the anonymous reviewers of ACL for their valuable feedback. C.A. is supported, in part, by grants from Capital One, LaCross Institute for Ethical AI in Business, the UVA Environmental Institute, OpenAI Researcher Program, Thinking Machine's Tinker Research Grant, and Cohere. The views expressed are those of the authors and do not reflect the official policy or the position of the funding agencies.\\

\bibliography{custom}

\appendix



\appendix

\raggedbottom

\part{Appendix}
\parttoc

\section{LLM-as-a-Judge Verification Protocol}
\label{sec:appendix}

Inspired by \citep{chen2024huatuogpt}, We employ an LLM-as-a-judge framework to automatically evaluate the correctness of model-generated responses. In this setup, GPT-4o acts as a verifier that compares a model’s prediction against a reference answer and determines whether the response is logically correct and linguistically valid. The verifier outputs a binary decision, returning \texttt{True} when the response aligns with the reference and \texttt{False} otherwise. Fig.~\ref{fig:judge_prompt} shows the exact prompt used for verification.

\begin{figure}[h]
\centering
\small
\begin{tcolorbox}[
  width=\columnwidth,
  colback=gray!5,
  colframe=black,
  title=\textbf{Prompt for the LLM-as-a-Judge Evaluator},
  fonttitle=\bfseries,
  top=1mm,
  bottom=1mm,
  left=1.5mm,
  right=1.5mm,
  boxrule=0.6pt
]

\texttt{<Model Response>}\\
\textcolor{blue!70!black}{\texttt{\{Model Response\}}}\\
\texttt{</Model Response>}\\[1.5mm]

\texttt{<Reference Answer>}\\
\textcolor{red!80!black}{\texttt{\{Ground-truth Answer\}}}\\
\texttt{</Reference Answer>}\\[2mm]

You are given a model-generated response and a reference answer. 
Determine whether the model response is correct with respect to the reference.
Output \texttt{"True"} if the response is correct and \texttt{"False"} otherwise.

\end{tcolorbox}
\caption{Prompt used for LLM-as-a-judge verification.}
\label{fig:judge_prompt}
\end{figure}

\section{Training and Verification Protocols}
\label{sec:training_verification}

This section documents the prompts, reward verification procedures, and training hyperparameters used for supervised and reinforcement fine-tuning. Together, these components define the optimization signals and structured supervision underlying the proposed framework.

\subsection{Reward Verification and Weighting}
\label{sec:appendix-reward}

We design a composite reward that jointly enforces clinical correctness, language fidelity, and output format compliance. The final reward is defined as
\[
R = 0.65 \times R_{\text{accuracy}} + 0.30 \times R_{\text{language}} + 0.05 \times R_{\text{format}}.
\]
This weighting prioritizes medical correctness while explicitly penalizing language drift and format violations.

\subsection{Verifier Models and Prompts}

Both correctness and language rewards are scored using \textbf{gpt-4.1} with \texttt{temperature=0.0} and \texttt{max\_tokens=10}. For each prompt, we generate 16 candidate responses to estimate stable reward signals.

\subsection{Accuracy Verifier.}
\begin{tcolorbox}[colback=gray!5!white, colframe=gray!70, boxrule=0.5pt, sharp corners, left=4pt, right=4pt, top=4pt, bottom=4pt]
\small
You are an expert multilingual medical evaluator. Score the generated response for correctness and medical validity on a continuous scale from 0.0 to 1.0. Give 1.0 if the reasoning is clinically sound and semantically correct, even if phrased differently from the reference. Focus on factual and clinical accuracy rather than wording.

Question: \{question\}

Ground truth answer: \{ground\_{}truth\}


Generated response: \{generated\}

Output only a float between 0.0 and 1.0.
\end{tcolorbox}

\subsubsection{Language Consistency Verifier Prompt} 

\begin{tcolorbox}[colback=gray!5!white, colframe=gray!70, boxrule=0.5pt, sharp corners, left=4pt, right=4pt, top=4pt, bottom=4pt]
\small
You are an expert multilingual medical evaluator. Determine whether the model response is written entirely in the same language as the question.

Question language: \{language\}

Generated response: \{generated\}

Output 1.0 if the language matches exactly; otherwise output 0.0.
\end{tcolorbox}

\subsubsection{Format Reward}
We apply a deterministic rule-based check requiring exactly one \texttt{<thinking>} block and one \texttt{<answer>} block, implemented using regular expressions with \texttt{re.DOTALL}. This constraint ensures consistent structure during reinforcement learning.

\subsection{Training Hyperparameters}
\label{app:training-hparams}

\paragraph{Compute and reproducibility configuration.}
All experiments were run on a single node equipped with $8\times$ NVIDIA A100 GPUs (80GB each). We used distributed training via DeepSpeed ZeRO-3 across 8 GPUs. The software stack was PyTorch 2.1, Transformers 4.40, DeepSpeed 0.14, and CUDA 12.1. For reproducibility, we report results over 3 independent runs and fix the random seed to 42.

\paragraph{Compute budget.}
Wall-clock training time was approximately 6--12 hours for SFT and 10--20 hours for GRPO-based reinforcement fine-tuning, corresponding to roughly 130--250 total GPU-hours.

\subsubsection{Supervised Fine-Tuning (SFT).}
\begin{itemize}
\item Optimizer: AdamW ($\beta_1=0.9$, $\beta_2=0.999$)
\item Learning rate: $1 \times 10^{-5}$ (cosine scheduler, warmup ratio 0.1)
\item Epochs: 3
\item Effective batch size: 32
\item Max sequence length: 4096
\item Precision: bf16
\item Distributed optimization: DeepSpeed ZeRO-3 (single-node, 8 GPUs) with gradient checkpointing
\end{itemize}

\subsection{SFT and RFT Prompts.}
\label{app:sft-rft-prompts}

As shown in Figure~\ref{fig:sft-rft-prompt}, we adopt the instruction and formatting template from \citet{hwang2025learn} with minor modifications for our multilingual medical reasoning setting. We use the same prompt structure for both supervised fine-tuning (SFT) and reinforcement fine-tuning (RFT); the only difference lies in the training objective, not in the prompt text.

\begin{figure}[h]
\begin{tcolorbox}[colback=gray!5!white, colframe=gray!70, boxrule=0.5pt, sharp corners, left=4pt, right=4pt, top=4pt, bottom=4pt]
\small
\textbf{System message 1.} You are an expert multilingual medical doctor. When answering a medical question, follow these steps:
\begin{enumerate}
    \item First, search your internal knowledge base thoroughly for relevant background information about the topic.
    \item Understand and reason the question fully in English first.
    \item Reason mainly in English, but code-switch naturally into the target language whenever useful for clarity or domain accuracy.
    \item Consider multiple perspectives and potential answers before settling on your final response.
    \item Evaluate the confidence in your answer based on the information available to you.
    \item Provide the final answer clearly in the target language, making sure it's well-supported by your reasoning.
    \item If there are significant uncertainties or gaps in your knowledge, acknowledge them transparently.
\end{enumerate}
Your goal is to provide accurate, well-reasoned responses that demonstrate depth of understanding, not just surface-level answers.

\vspace{6pt}
\textbf{System message 2.} You are an expert multilingual medical doctor. When answering a medical question, think and reason mainly in English with natural code-switching to the target language. Use multi-step reasoning wrapped in \texttt{<step>} tags inside \texttt{<thinking>}.

\vspace{6pt}
\textbf{User message.} The question is in \{language\}. \{question\}
Please think carefully with English-guided reasoning and code-switching, return your reasoning inside \texttt{<thinking>} \texttt{</thinking>} tags, and the final answer inside \texttt{<answer>} \texttt{</answer>} tags. Final answer ONLY in \{language\}.

\end{tcolorbox}
\caption{SFT and RFT prompt template used for multilingual medical reasoning.}
\label{fig:sft-rft-prompt}
\end{figure}





\newpage

\subsubsection{Reinforcement Fine-Tuning (RFT).}
\begin{itemize}
\item Algorithm: GRPO
\item Learning rate: $1 \times 10^{-6}$ (cosine scheduler, warmup ratio 0.1)
\item Weight decay: 0.1
\item Effective batch size: 16
\item Generations per prompt: 16
\item Max training steps: 500
\item Max prompt / completion length: 1024 / 1024
\item Distributed optimization: DeepSpeed ZeRO-3 (single-node, 8 GPUs)
\end{itemize}

\section{Dataset Details}
\label{sec:dataset_composition}

\begin{figure*}[!t]
\centering
\setlength{\tabcolsep}{6pt}
\begin{minipage}[h]{0.48\textwidth}
\vspace{0pt}
\centering
\includegraphics[width=\linewidth]{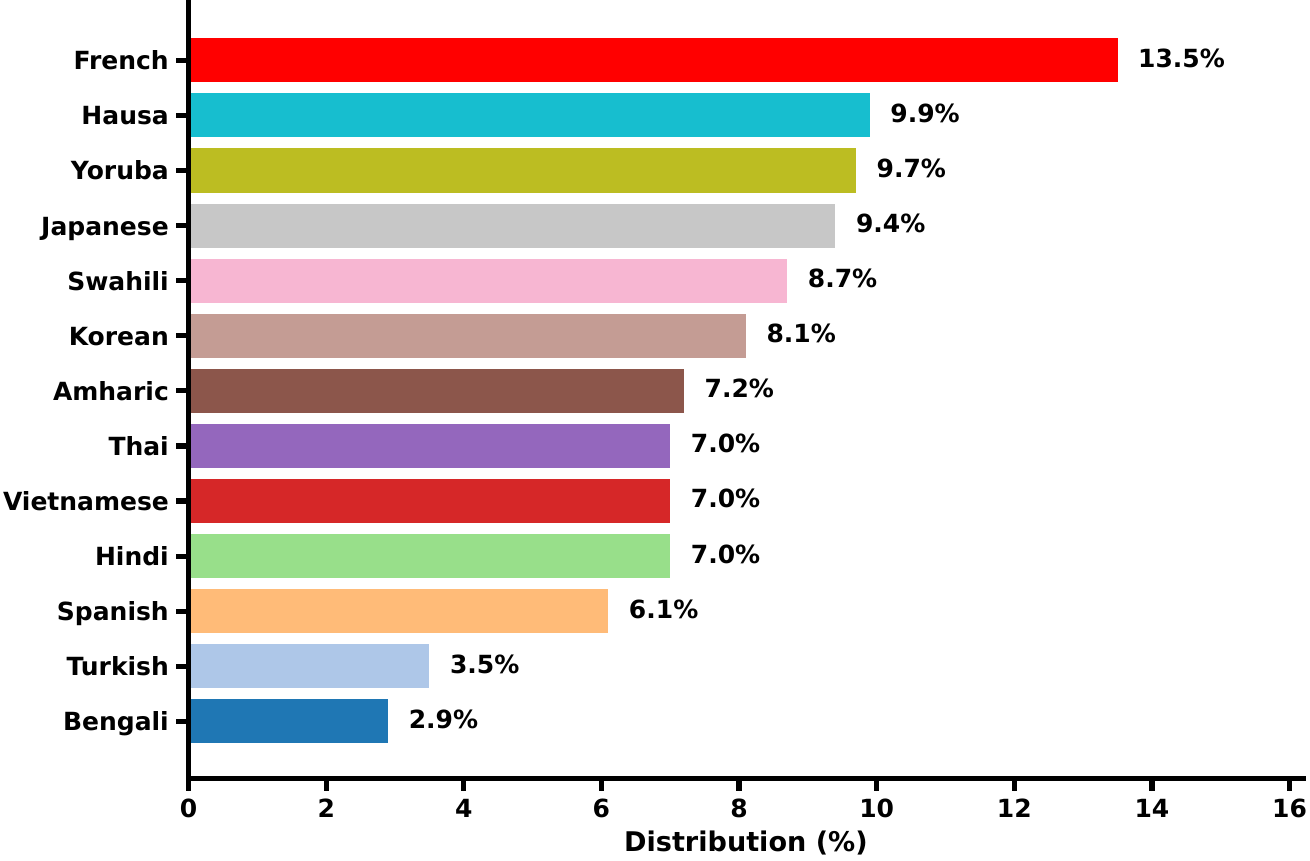}
\end{minipage}
\hfill
\begin{minipage}[h]{0.48\textwidth}
\vspace{10pt} 
\centering
\renewcommand{\arraystretch}{1.2} 
\footnotesize 
\begin{tabularx}{\linewidth}{>{\raggedright\arraybackslash}p{0.35\linewidth} >{\raggedright\arraybackslash}X}
\toprule
\textbf{Language Family} & \textbf{Languages} \\
\midrule
\rowcolor{pink!25}
Afroasiatic & Amharic (Am), Hausa (Ha) \\
\rowcolor{yellow!25}
Niger--Congo & Swahili (Sw), Yoruba (Yo) \\
\rowcolor{green!25}
Indo-European & Bengali (Bn), Hindi (Hi), French (Fr), Spanish (Es) \\
\rowcolor{cyan!20}
Turkic & Turkish (Tr) \\
\rowcolor{orange!20}
Austroasiatic & Vietnamese (Vi) \\
\rowcolor{blue!20}
Tai--Kadai & Thai (Th) \\
\rowcolor{red!20}
Japonic & Japanese (Ja) \\
\rowcolor{purple!20}
Koreanic & Korean (Ko) \\
\bottomrule
\end{tabularx}
\end{minipage}
\caption{\textbf{Language and family composition of \textsc{CureMed-Bench}.} \textbf{Left:} Number of dataset instances per language across the 13 languages. \textbf{Right:} Assignment of languages to eight language families with standard abbreviations.}

\label{fig:dataset_overview}
\end{figure*}

This appendix characterizes the linguistic composition of \textsc{CureMed-Bench}. Figure~\ref{fig:dataset_overview} shows the per-language instance distribution, with French contributing the largest share (13.5\%) and Bengali the smallest (2.9\%), and most languages occupying a mid-range band of roughly 7--10\% of the data. The figure also groups the 13 languages into eight language families, spanning Afroasiatic and Niger--Congo as well as Indo--European, Turkic, Austroasiatic, Tai--Kadai, Japonic, and Koreanic. Together, these statistics highlight both the dataset’s uneven language coverage and its broad typological diversity.

\subsection{Language-based Curriculum Tiers}
\label{sec:curriculum_language_tiers}

We construct our curriculum by defining difficulty along the linguistic axis rather than by question complexity. To operationalize this design, we use Qwen2.5-14B-Instruct as a reference model and estimate baseline reasoning accuracy separately for each language. The model performs best on high-resource languages and degrades as linguistic resources and model familiarity decrease, so we treat high-resource languages as easier tasks and progressively introduce more challenging languages during training. This curriculum aims to transfer reasoning competence learned in high-resource settings to underrepresented languages while maintaining language fidelity.

Based on the baseline accuracy ranking, we partition languages into three tiers. The high-resource tier includes French, Japanese, Spanish, and Vietnamese. The medium-resource tier includes Korean, Thai, Turkish, and Bengali. The low-resource tier includes Amharic, Yoruba, Hausa, Hindi, and Swahili. This tiering reflects the reference model’s initial proficiency distribution and provides a structured progression from easier to harder multilingual reasoning conditions.

\section{Data Curation}
\label{app:dataset_details}
The following prompt was used to generate the initial pool of medically grounded multiple-choice questions across 13 languages. Inspired by the  approach of \citet{hwang2025learn} and \citet{zhang2305huatuogpt}, we adapted their template and instructed GPT-4o to query MedlinePlus directly and independently construct questions in each target language rather than translating from a shared source. This ensures linguistic naturalness, cultural appropriateness, and strong 
domain grounding across all languages.

\subsection{Why an MCQ-to-Open-Ended Pipeline?}
\label{app:mcq_to_openended}
Following prior medical LLM data curation workflows, we adopt a two-step MCQ-to-open-ended pipeline: (i) generate and filter source-grounded multiple-choice questions (MCQs), then (ii) convert the retained items into open-ended queries \citep{hwang2025learn,zhang2305huatuogpt}. We prefer this over direct open-ended generation for three reasons. First, MCQs are easier to verify at scale because the correct answer is discrete, enabling efficient automated checks and filtering. Second, MCQ structure provides an explicit anchor that helps reduce ambiguity and the risk of ungrounded content (hallucinations) during generation. Third, this approach improves practicality in multilingual settings by strengthening quality control before producing open-ended questions that better reflect real use while remaining strictly evaluable \citep{hwang2025learn,zhang2305huatuogpt}.

\subsection{Construction of the SFT Dataset}
\label{app:sft_dataset_construction}

For each $(\textit{question}, \textit{target language}, \textit{gold answer})$ triplet in our curated set, we use a strong teacher model (GPT-4o) to produce supervised fine-tuning targets. Specifically, the teacher is prompted to generate: (i) a stepwise reasoning trace that may mix English clinical terminology with the target language for precision and readability, and (ii) a final answer that must be written entirely in the target language. To standardize training targets across languages, the teacher output is constrained to a fixed schema consisting of a \texttt{<thinking>} block containing explicit \texttt{<step>} tags, followed by an \texttt{<answer>} block containing only the final answer.

After generation, we apply human verification to ensure (1) the final answer is clinically consistent with the gold reference answer, and (2) the output satisfies formatting and language constraints, most importantly, that the content inside \texttt{<answer>} is strictly in the target language. Samples that pass verification are retained as SFT training instances. The resulting SFT dataset is then split according to the protocol described in Section~4.1 (\S\ref{exp:data_splits}).

\begin{figure*}[h]
\centering
\begin{tcolorbox}[
  colback=gray!12,
  colframe=black,
  boxrule=0.6pt,
  arc=2pt,
  left=6pt,right=6pt,top=5pt,bottom=5pt,
  width=\textwidth
]
\footnotesize
\textbf{Prompt for Generating Multilingual Medical Multiple-Choice Questions}

\vspace{2pt}
\textbf{Task:} You are an expert medical content generator. Generate \{num\_questions\} high-quality, medically accurate multiple-choice questions (MCQs) based strictly on content from MedlinePlus by searching and curating from the website.

\vspace{2pt}
You must independently compose each question in \textbf{ALL} of the following languages: Amharic, Bengali, French, Hausa, Hindi, Japanese, Korean, Spanish, Swahili, Thai, Turkish, Vietnamese, Yoruba.

\vspace{2pt}
\textbf{Requirements:}
\begin{enumerate}
  \setlength{\itemsep}{1pt}
  \setlength{\topsep}{1pt}
  \setlength{\parsep}{0pt}
  \setlength{\parskip}{0pt}
  \item \textbf{Medical Grounding:} All information must be sourced from MedlinePlus, covering symptoms, causes, risk factors, diagnostics, treatments, or prevention strategies.
  \item \textbf{Independent Composition:} Each language version must be originally written (not translated) using natural phrasing and medically appropriate terminology for that language.
  \item \textbf{Clinical Reasoning Depth:} Questions must require genuine clinical reasoning beyond trivial fact recall. Each question should have exactly one unambiguous correct answer.
  \item \textbf{Format:} 4-option MCQ (A/B/C/D) with one correct answer.
\end{enumerate}

\vspace{2pt}
\textbf{Output Format:} Return valid JSON array:
\vspace{2pt}

{\ttfamily\scriptsize\raggedright
\detokenize{[}\par
\detokenize{  \{"question_id": "<id>", "source_concept": "<MedlinePlus_topic>",}\par
\detokenize{   "mcq_items": [\{"language_code": "<lang>", "question": "<text>",}\par
\detokenize{     "option_A": "<text>", "option_B": "<text>", "option_C": "<text>", "option_D": "<text>",}\par
\detokenize{     "correct_answer": "<A|B|C|D>"\}, ...]\}}\par
\detokenize{]}\par
}

\vspace{2pt}
\textbf{IMPORTANT:} Return ONLY valid JSON without explanations, formatting, or additional text. Ensure all special characters are properly escaped.
\end{tcolorbox}

\caption{Prompt for Stage~1 multilingual MCQ generation. Here, \texttt{\{num\_questions\}} specifies the number of questions to generate, and GPT-4o queries MedlinePlus directly to construct clinically grounded questions independently in each of the 13 target languages.}
\label{fig:mcq-generation-prompt}
\end{figure*}

\subsection{Human Verification Protocol and Rater Instructions}
\label{sec:human_verification_instructions}

This section documents the human verification procedures used to validate the quality of our synthetic data. Participation in the study was completely voluntary, as participants received no payment for participating in the study. We provide the exact instructions used by medical professionals who assessed the clinical correctness of question-answer pairs and by native speakers who evaluated the language's correctness and fidelity in the target language, as shown in Figures \ref{fig:med_verification_instructions} and \ref{fig:native_language_verification_instructions}. These materials specify the task setup, scoring rubric, and optional comment guidelines used throughout our verification pipeline.


\begin{figure*}[!t]
\centering
\small
\begin{tcolorbox}[
  colback=gray!12,
  colframe=black,
  boxrule=0.6pt,
  arc=2pt,
  left=6pt,right=6pt,top=6pt,bottom=6pt,
  width=\linewidth
]
{\bfseries Participant Instructions: Verification Task}

\vspace{6pt}
{\bfseries Task Overview}\par
You will review synthetically generated medical question--answer pairs based on public sources such as MedlinePlus. These pairs are generated synthetically and do not involve real patient data. Your role is to assess medical correctness and accuracy.

\vspace{6pt}
{\bfseries What You Will Do}\par
For each question--answer pair:
\begin{itemize}\setlength{\itemsep}{2pt}\setlength{\parskip}{0pt}
  \item Read the question and the provided answer.
  \item Check for medical correctness: ensure the information is accurate, logically sound, and aligned with standard medical knowledge.
  \item Assign a score from 1 to 5:
  \begin{itemize}\setlength{\itemsep}{1pt}\setlength{\parskip}{0pt}
    \item \textbf{1}: Completely inaccurate or misleading.
    \item \textbf{2}: Mostly inaccurate with major errors.
    \item \textbf{3}: Partially accurate but with notable issues.
    \item \textbf{4}: Mostly accurate with minor issues.
    \item \textbf{5}: Fully accurate and reliable.
  \end{itemize}
  \item (Optional) Provide a brief comment if necessary (e.g., explain errors, suggest corrections, or note cultural/language specifics). Comments are optional but helpful.
\end{itemize}

\vspace{2pt}
You will receive batches of 50--100 pairs via an online survey. The task takes approximately 1--2 hours and can be completed remotely at your convenience. You may skip any pair or stop at any time.

\end{tcolorbox}
\caption{Instructions provided to medical professional annotators for verifying clinical correctness of synthetic question--answer pairs.}
\label{fig:med_verification_instructions}
\end{figure*}


\begin{figure*}[!t]
\centering
\small
\begin{tcolorbox}[
  colback=gray!12,
  colframe=black,
  boxrule=0.6pt,
  arc=2pt,
  left=6pt,right=6pt,top=6pt,bottom=6pt,
  width=\linewidth
]
{\bfseries Participant Instructions: Language Verification Task}

\vspace{6pt}
{\bfseries Task Overview}\par
You will review synthetically generated medical question--answer pairs written in one of the following target languages: Amharic, Bengali, French, Hausa, Hindi, Japanese, Korean, Spanish, Swahili, Thai, Turkish, Vietnamese, and Yoruba. These pairs are generated synthetically and do not include real patient data. Your role is to verify whether the question and answer are written correctly and naturally in the target language.

\vspace{6pt}
{\bfseries What You Will Do}\par
For each question--answer pair:
\begin{itemize}\setlength{\itemsep}{2pt}\setlength{\parskip}{0pt}
  \item Read the question and the provided answer.
  \item Verify language correctness and fidelity:
  \begin{itemize}\setlength{\itemsep}{1pt}\setlength{\parskip}{0pt}
    \item The text is in the requested target language (no switching to another language).
    \item The wording is grammatical and understandable for a native speaker.
    \item The phrasing is natural and appropriate for medical communication.
    \item Medical terms are expressed in an acceptable way for the target language (including common loanwords, when appropriate).
  \end{itemize}
  \item Assign a score from 1 to 5:
  \begin{itemize}\setlength{\itemsep}{1pt}\setlength{\parskip}{0pt}
    \item \textbf{1}: Not in the target language or largely unintelligible.
    \item \textbf{2}: Major language errors; difficult to understand.
    \item \textbf{3}: Understandable but with noticeable errors or unnatural phrasing.
    \item \textbf{4}: Mostly correct and natural with minor issues.
    \item \textbf{5}: Fully correct, natural, and clearly in the target language.
  \end{itemize}
  \item (Optional) Provide a brief comment to note issues (e.g., incorrect language, grammar problems, unnatural phrasing, or better word choices).
\end{itemize}

\vspace{2pt}
You will receive batches of 50--100 pairs via an online survey. The task takes approximately 1--2 hours and can be completed remotely at your convenience. You may skip any pair or stop at any time.
\end{tcolorbox}
\caption{Instructions provided to native-speaker annotators for verifying language correctness and target-language fidelity of synthetic question--answer pairs.}
\label{fig:native_language_verification_instructions}
\end{figure*}


\subsection{Human Verification Scores by Language}
\label{sec:human_scores_by_language}
\looseness=-1 We report per-language human verification scores from two rater groups. Medical professionals score clinical correctness of each question--answer pair, while native speakers score target-language quality and fidelity. Table~\ref{tab:human_verification_scores} summarizes both scores on a 1--5 scale, where higher values indicate better quality. Scores are generally high across languages, reflecting the rigorous curation process applied during dataset construction. Notably, lower-resource languages such as Amharic, Thai, and Turkish receive slightly lower scores, suggesting that these languages present greater challenges for both clinical accuracy and fluent translation. Overall, the consistently high ratings across both dimensions confirm that the dataset meets the quality bar required for reliable multilingual medical reasoning evaluation.

\begin{table}[h]
\centering
\small
\setlength{\tabcolsep}{6pt}
\renewcommand{\arraystretch}{1.08}
\begin{tabular}{l c c}
\toprule
\textbf{Language} & \textbf{Medical correctness} & \textbf{Language quality} \\
\midrule
Amharic    & 4.45 & 4.45 \\
Bengali    & 4.92 & 4.96 \\
French     & 5.00 & 5.00 \\
Hausa      & 4.96 & 5.00 \\
Hindi      & 5.00 & 4.92 \\
Japanese   & 4.96 & 4.96 \\
Korean     & 5.00 & 5.00 \\
Spanish    & 5.00 & 5.00 \\
Swahili    & 5.00 & 4.96 \\
Thai       & 4.70 & 4.70 \\
Turkish    & 4.60 & 4.60 \\
Vietnamese & 4.95 & 4.95 \\
Yoruba     & 5.00 & 5.00 \\
\bottomrule
\end{tabular}
\caption{Per-language human verification scores (1--5) from medical professionals (clinical correctness) and native speakers (language quality). Higher is better.}
\label{tab:human_verification_scores}
\end{table}

\section{Per-Language Model Performance}
\label{sec:per_language}

This section provides a fine-grained analysis of multilingual medical reasoning performance broken down by language. We compare \method with instruction-tuned baselines across all 13 languages in \methodBench, enabling a detailed examination of logical correctness and language consistency under diverse linguistic and resource conditions. This per-language view complements aggregate results by revealing where gains are most pronounced and where challenges remain.

\subsection{Per-Language Results for Qwen2.5-7B}
\label{sec:per_language_7b}
\looseness=-1 Table~\ref{tab:per_language_7b_base_vs_curemed} reports per-language performance for the Qwen2.5-7B-Instruct baseline and its \method variant. Across all 13 languages, \method substantially improves both logical accuracy and language consistency. Gains are especially large in low-resource languages such as Amharic, Hausa, Swahili, and Yoruba, where the baseline frequently fails to produce correct or language-faithful responses. In higher-resource languages such as French, Japanese, and Spanish, \method yields more moderate but consistent improvements, indicating that GRPO-guided curriculum RL enhances reasoning robustness without degrading performance in well-resourced settings. Overall, these results show that \method improves multilingual medical reasoning uniformly while significantly narrowing performance disparities across languages.

\begin{table*}[h]
\centering
\small
\setlength{\tabcolsep}{1.2pt}
\renewcommand{\arraystretch}{0.9}

\definecolor{CUREgreen}{RGB}{226,239,218}

\begin{tabular}{l c >{\columncolor{CUREgreen}}c c c >{\columncolor{CUREgreen}}c c}
\toprule

\textbf{Language} &
Logic (Base) &
Logic (\method) &
$\Delta$ &
Lang. (Base) &
Lang. (\method) &
$\Delta$ \\
\midrule

Amharic    & 0.95  & 17.14 & \textbf{+16.19} & 0.00  & 64.76 & \textbf{+64.76} \\
Bengali    & 10.00 & 60.00 & \textbf{+50.00} & 2.14  & 91.43 & \textbf{+89.29} \\
French     & 67.86 & 77.86 & \textbf{+10.00}  & 71.43 & 96.43 & \textbf{+25.00} \\
Hausa      & 5.06  & 43.04 & \textbf{+37.98} & 0.00  & 77.22 & \textbf{+77.22} \\
Hindi      & 4.48  & 48.51 & \textbf{+44.03} & 5.97  & 90.30 & \textbf{+84.33} \\
Japanese   & 68.57 & 77.14 & \textbf{+8.57}  & 60.00 & 94.29 & \textbf{+34.29} \\
Korean     & 41.33 & 52.00 & \textbf{+10.67} & 26.67 & 84.00 & \textbf{+57.33} \\
Spanish    & 62.86 & 72.38 & \textbf{+9.52}  & 60.95 & 96.19 & \textbf{+35.24} \\
Swahili    & 0.00  & 35.71 & \textbf{+35.71} & 0.00  & 67.14 & \textbf{+67.14} \\
Thai       & 51.02 & 59.18 & \textbf{+8.16}  & 37.76 & 86.73 & \textbf{+48.97} \\
Turkish    & 12.50 & 43.75 & \textbf{+31.25} & 3.57  & 75.89 & \textbf{+72.32} \\
Vietnamese & 66.67 & 70.48 & \textbf{+3.81}  & 61.90 & 94.29 & \textbf{+32.39} \\
Yoruba     & 0.00  & 40.86 & \textbf{+40.86} & 0.00  & 77.42 & \textbf{+77.42} \\

\bottomrule
\end{tabular}
\vspace{-0.1in}
\caption{
Per-language performance of Qwen2.5-7B-Instruct (Base) and the \method 7B variant on \textsc{CureMed-Bench}. We report logical correctness and language accuracy, along with absolute gains $\Delta$ (\method$-$Base).
}
\label{tab:per_language_7b_base_vs_curemed}
\end{table*}

\subsection{Per-Language Results for Qwen2.5-3B}
\label{sec:per_language_3b}

\looseness=-1 Table~\ref{tab:per_language_3b_base_vs_curemed} shows that \method consistently improves the 3B model across all evaluated languages in both logical correctness and language accuracy. The baseline 3B model exhibits extremely low performance for several languages, including Amharic, Hausa, Swahili, and Turkish, whereas the \method variant achieves large absolute gains, often exceeding 40--80 percentage points. Even in languages where the base model is already relatively stronger, such as French, Japanese, Spanish, and Vietnamese, \method delivers clear and reliable improvements. These results demonstrate that curriculum-guided reinforcement is particularly effective for small models, enabling robust multilingual medical reasoning despite limited model capacity.

\begin{table*}[h]
\centering
\small
\setlength{\tabcolsep}{1.2pt}
\renewcommand{\arraystretch}{0.9}
\definecolor{CUREgreen}{RGB}{226,239,218}

\begin{tabular}{l c >{\columncolor{CUREgreen}}c c c >{\columncolor{CUREgreen}}c c}
\toprule

\textbf{Language} &
Logic (Base) &
Logic (\method) &
$\Delta$ &
Lang. (Base) &
Lang. (\method) &
$\Delta$ \\
\midrule

Amharic    & 0.95  & 14.29 & \textbf{+13.34} & 0.00  & 40.95 & \textbf{+40.95} \\
Bengali    & 2.86  & 55.71 & \textbf{+52.85} & 0.00  & 85.00 & \textbf{+85.00} \\
French     & 12.14 & 70.71 & \textbf{+58.57} & 22.14 & 95.71 & \textbf{+73.57} \\
Hausa      & 2.53  & 27.85 & \textbf{+25.32} & 0.00  & 64.56 & \textbf{+64.56} \\
Hindi      & 5.97  & 28.36 & \textbf{+22.39} & 0.00  & 83.58 & \textbf{+83.58} \\
Japanese   & 23.81 & 62.86 & \textbf{+39.05} & 26.67 & 89.52 & \textbf{+62.85} \\
Korean     & 8.00  & 36.00 & \textbf{+28.00} & 2.67  & 76.00 & \textbf{+73.33} \\
Spanish    & 17.14 & 62.86 & \textbf{+45.72} & 23.81 & 94.29 & \textbf{+70.48} \\
Swahili    & 0.00  & 17.86 & \textbf{+17.86} & 0.00  & 51.43 & \textbf{+51.43} \\
Thai       & 10.20 & 58.16 & \textbf{+47.96} & 0.00  & 73.47 & \textbf{+73.47} \\
Turkish    & 1.79  & 28.57 & \textbf{+26.78} & 0.00  & 53.57 & \textbf{+53.57} \\
Vietnamese & 44.76 & 69.52 & \textbf{+24.76} & 79.05 & 80.00 & \textbf{+0.95} \\
Yoruba     & 6.45  & 17.20 & \textbf{+10.75} & 0.00  & 69.89 & \textbf{+69.89} \\

\bottomrule
\end{tabular}

\caption{
Per-language performance of the 3B Base model and its \method variant on \textsc{CureMed-Bench}. We report logical correctness and language accuracy, along with absolute gains $\Delta$ (\method$-$Base).
}\vspace{-0.1in}
\label{tab:per_language_3b_base_vs_curemed}
\end{table*}

\subsection{Proprietary Model Performance on \methodBench}
\label{sec:closed_source_summary}
Table~\ref{tab:closed-source-curemed-concise} summarizes inference-only performance of frontier models on \methodBench, reporting language consistency and logical accuracy averaged over 13 languages. While some models maintain strong target-language adherence (\eg Claude~3 Haiku), results reveal substantial brittleness: GPT-5-nano exhibits notably weaker language consistency, and the Gemini 2.5 family degrades sharply in both language control and reasoning quality (with Gemini 2.5 Pro nearly collapsing). These averages also conceal larger failures in low-resource languages, where models more frequently drift from the target language and show steeper drops in logical accuracy (see Appendix~\ref{sec:closed_source_model}). Overall, \methodBench exposes a reliability gap for proprietary LLMs: strong performance in some settings does not ensure robust multilingual reasoning or consistent target-language adherence.

\begin{table}[!t]
\centering
\small
\setlength{\tabcolsep}{1.2pt}
\renewcommand{\arraystretch}{0.9}
\begin{tabular}{lcc}
\toprule
\textbf{Model} &
\textbf{Lang. Consistency} ($\uparrow$) &
\textbf{Logical Acc.} ($\uparrow$) \\
\midrule
GPT-5-nano       & 69.11 & 73.24 \\
GPT-5-mini       & 75.33 & 80.57 \\
Gemini~2.5 Flash & 48.01 & 54.79 \\
Gemini~2.5 Pro   & 4.33  & 10.62 \\
Claude~3 Haiku   & 93.43 & 73.31 \\
\bottomrule
\end{tabular}
\caption{Inference-only performance of proprietary models on \methodBench (averaged across 13 languages).}
\vspace{-0.1in}
\label{tab:closed-source-curemed-concise}
\end{table}

\subsection{Per-language Performance of Closed-source Models}
\label{sec:closed_source_model}

We report per language results for proprietary models on \methodBench using logical accuracy (Table~\ref{tab:closed-source-logical-by-language}) and language consistency (Table~\ref{tab:closed-source-lang-consistency-by-language}). Although aggregate scores are strong for several systems, per language analysis reveals clear cross lingual brittleness. Higher resource languages such as French and Spanish show consistently high logical accuracy and strong target language adherence, with similarly stable behavior in Japanese, Korean, Thai, Turkish, and Vietnamese. In contrast, low resource languages expose systematic failures. Amharic shows frequent target language breakdown for several models, where language consistency drops sharply even when accuracy remains non trivial for some settings. Hausa exhibits a different pattern in which models drift from the target language despite moderate to high logical accuracy, indicating that medical reasoning does not guarantee language control under inference only prompting. Yoruba is the most challenging overall, with both language consistency and logical accuracy decreasing across models. Overall, these results motivate evaluating multilingual medical reasoning using joint measures of correctness and target language fidelity.

\section{Failure Mode Analysis}

To move beyond aggregate leaderboard metrics, we analyze the failure patterns of baseline models and ablation variants on \textsc{CureMed-Bench}. Our framework evaluates logical accuracy and language consistency independently on open ended questions with a single verifiable answer, revealing three recurring failure modes that motivate \method. These failure modes are not mutually exclusive; a single model may exhibit several simultaneously, and their co-occurrence often masks the true source of performance degradation under standard evaluation. Understanding each failure mode in isolation allows us to trace specific architectural and training choices to their downstream effects, and to design targeted interventions that address the root causes rather than surface level symptoms.

\subsection*{Failure Mode I: Clinically Plausible but Logically Incorrect Reasoning}

The most common failure occurs when models produce fluent and well structured responses that contain incorrect clinical reasoning. Figure~\ref{fig:curemed-spanish-sidebyside-newq} illustrates this pattern. The Qwen2.5-7B-Instruct baseline generates a coherent Spanish response but predicts an incorrect diagnosis, interpreting a mild viral presentation as early pulmonary infection despite the absence of fever and respiratory distress. Here, the model matches superficial symptom patterns instead of ruling out alternatives through differential reasoning, a behavior also noted in prior studies of medically unsound but persuasive model explanations~\cite{amann2020explainability, nori2023capabilities}.

\textsc{CureMed-Bench} exposes this failure more clearly than multiple choice benchmarks because it requires free form reasoning and a verifiable final answer. We observe severe reasoning errors even in domain specific models. In Table~\ref{tab:multilingual-medical-main}, \textsc{MedAlpaca-7B} reaches only 2.47\% logical accuracy and 3.50\% language consistency, while \textsc{Meditron-7B} reaches 2.50\% logical accuracy and 0.43\% language consistency. These results show that domain specific pretraining alone does not ensure reliable clinical reasoning under open ended evaluation.

\subsection*{Failure Mode II: Language Drift and Target Language Infidelity}

A second failure mode appears when models do not answer in the language of the query. They often switch to English or produce mixed language outputs. This tendency is consistent with the English dominant behavior reported in multilingual language models~\cite{nguyen2023democratizing, cahyawijaya2024llms}, especially in low resource languages.

Because \textsc{CureMed-Bench} evaluates logical accuracy and language consistency separately, it distinguishes reasoning ability from language fidelity and reveals a clear mismatch between them. For example, \textsc{OpenBioLLM-LLaMA3-8B} attains 36.62\% logical accuracy but only 1.47\% language consistency in Table~\ref{tab:multilingual-medical-main}. The model often reaches the correct medical conclusion but fails to express it in the required target language. Table~\ref{tab:closed-source-lang-consistency-by-language} shows that this problem persists even in strong proprietary systems. Gemini 2.5 Pro achieves only 1.90\% language consistency in Amharic and 2.15\% in Yoruba, despite showing nontrivial logical accuracy in those languages in Table~\ref{tab:closed-source-logical-by-language}.

These results show that reasoning correctness and target language fidelity do not emerge together by default. We therefore optimize both through the language consistency reward $R_{\text{lang}}$ and the code switching aware supervised fine tuning stage described in Sections~\ref{sec:cold} and~\ref{sec:reward-design}.

\subsection*{Failure Mode III: Systematic Low Resource Language Degradation}

The third failure mode is the sharp decline in performance for low resource languages. Table~\ref{tab:per_language_7b_base_vs_curemed} shows that Qwen2.5-7B-Instruct achieves 0.00\% language consistency and near zero logical accuracy in Amharic, Hausa, Swahili, and Yoruba, while performing much better in French with 71.43\% language consistency and 67.86\% logical accuracy, and in Japanese with 60.00\% language consistency and 68.57\% logical accuracy. This pattern points to weak cross lingual generalization rather than a general inability to perform medical reasoning.

Our ablations confirm that training design drives this failure. In Table~\ref{tab:cure_ablation}, naïve multilingual supervised fine tuning produces limited and sometimes negative gains, with logical accuracy dropping from 10.83\% to 9.50\% at the 3B scale. Table~\ref{tab:curriculum_order_half} provides stronger evidence: when training proceeds from low resource to high resource languages, performance degrades consistently across model sizes, and at the 7B scale, language consistency falls from 85.21\% to 64.21\%. These findings show that curriculum order strongly affects transfer to underrepresented languages. We address this failure with a high to low resource curriculum and retention aware data mixing ($\alpha = 0.85$), which establish stable early learning signals and support later transfer to lower resource settings.

\begin{table*}[h]
\centering
\small
\setlength{\tabcolsep}{3pt}
\renewcommand{\arraystretch}{1.03}

\begin{tabularx}{\textwidth}{l *{5}{>{\centering\arraybackslash}X}}
\toprule
\textbf{Language} &
\textbf{GPT-5-nano} &
\textbf{GPT-5-mini} &
\textbf{Gemini 2.5 Flash} &
\textbf{Gemini 2.5 Pro} &
\textbf{Claude 3 Haiku} \\
\midrule

Amharic     & 5.71  & 41.90 & 24.76 & 0.95  & 70.48 \\
Bengali     & 65.00 & 73.57 & 38.57 & 9.29  & 62.86 \\
French      & 89.29 & 93.57 & 75.71 & 23.57 & 90.71 \\
Hausa       & 78.48 & 89.87 & 43.04 & 3.80  & 55.70 \\
Hindi       & 78.36 & 78.36 & 62.69 & 14.93 & 76.12 \\
Japanese    & 84.76 & 84.76 & 69.52 & 13.33 & 87.62 \\
Korean      & 78.67 & 80.00 & 62.67 & 6.67  & 73.33 \\
Spanish     & 89.52 & 94.29 & 74.29 & 18.10 & 88.57 \\
Swahili     & 84.29 & 86.43 & 48.57 & 7.14  & 77.14 \\
Thai        & 85.71 & 90.82 & 64.29 & 6.12  & 75.51 \\
Turkish     & 79.46 & 84.82 & 53.57 & 6.25  & 70.54 \\
Vietnamese  & 88.57 & 88.57 & 63.81 & 18.10 & 84.76 \\
Yoruba      & 35.48 & 56.99 & 25.81 & 2.15  & 25.81 \\

\bottomrule
\end{tabularx}

\caption{
Logical accuracy (\%) of proprietary models on \methodBench across 13 languages under inference-only prompting. We report accuracy against the single ground-truth answer.
}
\label{tab:closed-source-logical-by-language}
\end{table*}

\begin{table*}[h]
\centering
\small
\setlength{\tabcolsep}{3pt}
\renewcommand{\arraystretch}{1.05}

\begin{tabularx}{\textwidth}{l *{5}{>{\centering\arraybackslash}X}}
\toprule
\textbf{Language} &
\textbf{GPT-5-nano} &
\textbf{GPT-5-mini} &
\textbf{Gemini 2.5 Flash} &
\textbf{Gemini 2.5 Pro} &
\textbf{Claude 3 Haiku} \\
\midrule

Amharic     & 1.90  & 24.76 & 12.38 & 1.90  & 92.38 \\
Bengali     & 39.29 & 65.71 & 34.29 & 1.43  & 95.71 \\
French      & 92.86 & 98.57 & 67.86 & 5.00  & 98.57 \\
Hausa       & 56.96 & 43.04 & 35.44 & 1.27  & 73.42 \\
Hindi       & 53.73 & 73.88 & 58.96 & 8.21  & 97.76 \\
Japanese    & 80.00 & 88.57 & 56.19 & 5.71  & 96.19 \\
Korean      & 92.00 & 88.00 & 56.00 & 4.00  & 97.33 \\
Spanish     & 97.14 & 98.10 & 71.43 & 5.71  & 91.43 \\
Swahili     & 82.86 & 77.86 & 32.86 & 2.86  & 98.57 \\
Thai        & 94.90 & 88.78 & 59.18 & 9.18  & 97.96 \\
Turkish     & 90.18 & 93.75 & 52.68 & 7.14  & 96.43 \\
Vietnamese  & 89.52 & 91.43 & 60.00 & 0.95  & 99.05 \\
Yoruba      & 27.96 & 32.26 & 23.66 & 2.15  & 67.74 \\

\bottomrule
\end{tabularx}

\caption{
Language consistency (\%) of proprietary models on \methodBench across 13 languages under inference-only prompting. We report the fraction of outputs that adhere to the requested target language.
}
\label{tab:closed-source-lang-consistency-by-language}
\end{table*}

\begin{table}
\centering
\small
\setlength{\tabcolsep}{3pt}
\renewcommand{\arraystretch}{1}

\definecolor{CUREgreen}{RGB}{226,239,218}
\definecolor{GroupBlue}{RGB}{225,235,245}

\begin{tabularx}{\linewidth}{l c *{3}{>{\centering\arraybackslash}X}}

\rowcolor{GroupBlue}
\textbf{Model} & \textbf{English} & \textbf{French} & \textbf{Italian} & \textbf{Spanish} \\

Qwen2.5-1.5B         & 1.40 & 6.40 & 4.80 & 6.80  \\
\rowcolor{CUREgreen}
~$\drsh$ \method     & \textbf{44.80} & \textbf{47.20} & \textbf{24.00} & \textbf{32.80} \\

Qwen2.5-3B           & 24.8  & 12.00  & 13.60 & 13.60  \\
\rowcolor{CUREgreen}
~$\drsh$ \method     & \textbf{48.00} & \textbf{50.60} & \textbf{36.80} & \textbf{48.80} \\

Qwen2.5-7B           & \textbf{54.40} & 44.00 & 34.40 & 48.00  \\
\rowcolor{CUREgreen}
~$\drsh$ \method     & 53.60 & \textbf{56.80} & \textbf{47.20} & \textbf{57.60} \\

Qwen2.5-14B          & 61.60 & 54.40 & 46.40 & 60.00  \\
\rowcolor{CUREgreen}
~$\drsh$ \method     & \textbf{66.40} & \textbf{64.40} & \textbf{64.80} & \textbf{68.00} \\

Qwen2.5-32B          & \textbf{72.80} & \textbf{73.60} & 64.80 & 70.40 \\
\rowcolor{CUREgreen}
~$\drsh$ \method     & 72.20 & 73.00 & \textbf{72.60} & \textbf{76.20} \\

\hline
\end{tabularx}

\caption{OOD accuracy on MedExpQA across four languages. \method improves reasoning performance across model sizes, showing cross-lingual generalization to unseen medical questions and languages.
}
\label{tab:crosslingual-exp-accuracy}
\end{table}

\begin{table}
\centering
\small
\setlength{\tabcolsep}{3pt}
\renewcommand{\arraystretch}{1}

\definecolor{CUREgreen}{RGB}{226,239,218}
\definecolor{GroupBlue}{RGB}{225,235,245}

\begin{tabularx}{\linewidth}{l *{3}{>{\centering\arraybackslash}X}}

\rowcolor{GroupBlue}
\textbf{Model} & \textbf{English} & \textbf{Simplified Chinese} & \textbf{Traditional Chinese} \\

Qwen2.5-1.5B         & 18.50 & 21.00  & 16.00 \\
\rowcolor{CUREgreen}
~$\drsh$ \method     & \textbf{37.80} & \textbf{59.50} & \textbf{47.50} \\

Qwen2.5-3B      & 32.50 & 55.00  & 36.00 \\
\rowcolor{CUREgreen}
~$\drsh$ \method     & \textbf{41.00} & \textbf{68.00} & \textbf{54.00} \\

Qwen2.5-7B           & 50.50 & \textbf{73.00} & \textbf{60.00} \\
\rowcolor{CUREgreen}
~$\drsh$ \method     & \textbf{51.50} & 70.00 & 57.00 \\

Qwen2.5-14B          & 56.00 & \textbf{80.50} & 69.50  \\
\rowcolor{CUREgreen}
~$\drsh$ \method     & \textbf{59.50} & 75.00 & \textbf{70.00} \\

Qwen2.5-32B          & 63.00 & \textbf{84.00} & 71.00 \\
\rowcolor{CUREgreen}
~$\drsh$ \method     & \textbf{64.00} & 81.00 & \textbf{76.00} \\

\hline
\end{tabularx}

\caption{OOD accuracy on MedQA across English and Chinese. \method improves reasoning performance across model sizes, demonstrating robustness across unseen languages.
}
\label{tab:crosslingualqa-accuracy-zh}
\end{table}

\section{Additional baselines}

\label{app:additional_baselines}

To strengthen our comparisons, we include additional modern baselines that cover general purpose distilled models (DeepSeek Qwen), recent multilingual language models (Qwen3), and medical reasoning focused models (MedReason, MedS3, and M1 variants). We report both \textit{Language Consistency} and \textit{Logical Accuracy} using the same evaluation protocol. 

Table~\ref{tab:additional_baselines_half} shows that performance generally increases with model scale within each baseline family. Across all comparisons, CURE MED achieves the strongest overall results, with particularly large gains in language consistency and consistent improvements in logical accuracy. The best performance is obtained by CURE MED Qwen2.5 32B, which reaches 94.96 language consistency and 70.04 logical accuracy, substantially exceeding the other baselines. 


\begin{table}[h]
\centering
\scriptsize
\setlength{\tabcolsep}{2.5pt}
\renewcommand{\arraystretch}{0.75}
\resizebox{\columnwidth}{!}{%
\begin{tabular}{lcc}
\toprule
\textbf{Model} & \textbf{Consistency ($\uparrow$)} & \textbf{Accuracy ($\uparrow$)} \\
\midrule
DeepSeek-Qwen 7B  & 1.12$\pm$0.31  & 28.65$\pm$0.84 \\
DeepSeek-Qwen 14B & 6.74$\pm$0.42  & 33.18$\pm$0.79 \\
DeepSeek-Qwen 32B & 11.83$\pm$0.58 & 40.27$\pm$0.74 \\
Qwen3-4B          & 24.18$\pm$0.95 & 22.76$\pm$0.88 \\
Qwen3-8B          & 33.96$\pm$0.90 & 31.24$\pm$0.82 \\
Qwen3-14B         & 43.71$\pm$0.86 & 38.62$\pm$0.77 \\
Qwen3-32B         & 55.83$\pm$0.80 & 48.57$\pm$0.71 \\
MedReason (8B)    & 26.74$\pm$0.89 & 34.12$\pm$0.81 \\
MedS3 (8B-PRM)    & 19.28$\pm$0.88 & 36.41$\pm$0.80 \\
m1-7B-23K         & 17.61$\pm$0.56 & 42.90$\pm$0.73 \\
m1-32B-1K         & 48.92$\pm$0.83 & 54.63$\pm$0.68 \\
\midrule
\textbf{CURE-MED} & & \\
CURE-MED-Qwen2.5-1.5B & 57.60$\pm$0.65 & 28.32$\pm$0.35 \\
CURE-MED-Qwen2.5-3B   & 74.28$\pm$0.60 & 42.93$\pm$0.60 \\
CURE-MED-Qwen2.5-7B   & 85.21$\pm$0.63 & 54.35$\pm$0.50 \\
CURE-MED-Qwen2.5-14B  & 90.27$\pm$0.31 & 63.74$\pm$0.43 \\
CURE-MED-Qwen2.5-32B  & \textbf{94.96$\pm$0.40} & \textbf{70.04$\pm$0.04} \\
\bottomrule
\end{tabular}%
}
\caption{Additional baseline results under the same evaluation protocol. We report mean$\pm$std over three runs for language consistency and logical accuracy.}
\label{tab:additional_baselines_half}
\end{table}


\section{Curriculum Ordering Ablation}
\label{app:curriculum_order}

We evaluate whether the ordering of our curriculum affects training outcomes. Our default schedule trains from high resource languages to medium resource languages and then to low resource languages. As a counterfactual, we reverse the schedule and train from low resource languages to medium resource languages and then to high resource languages. Due to resource constraints, we run this ablation for the 1.5B, 3B, and 7B settings only. Table~\ref{tab:curriculum_order_half} reports language consistency and logical accuracy under the same evaluation protocol.

The results show that curriculum ordering is important. Reversing the order yields a consistent degradation in both metrics across all model scales, indicating that the gains are not an artifact of a particular model size or random variation. At the same time, even under the reverse ordering, CURE MED remains stronger than representative size matched instruction tuned baselines, which suggests that the method is beneficial beyond the choice of schedule. Overall, these findings support the use of the proposed high resource to low resource ordering as it provides the most reliable improvements in both language consistency and logical accuracy.

\begin{table}[!htbp]
\centering
\scriptsize
\setlength{\tabcolsep}{2.5pt}
\renewcommand{\arraystretch}{0.85}
\resizebox{\columnwidth}{!}{%
\begin{tabular}{lcc}
\toprule
\textbf{Model} & \textbf{Consistency ($\uparrow$)} & \textbf{Accuracy ($\uparrow$)} \\
\midrule
\multicolumn{3}{l}{\textbf{Curriculum: High $\rightarrow$ Medium $\rightarrow$ Low}} \\
CURE-MED-Qwen2.5-1.5B & 57.60 $\pm$ 0.65 & 28.32 $\pm$ 0.35 \\
CURE-MED-Qwen2.5-3B   & 74.28 $\pm$ 0.60 & 42.93 $\pm$ 0.60 \\
CURE-MED-Qwen2.5-7B   & 85.21 $\pm$ 0.63 & 54.35 $\pm$ 0.50 \\
\midrule
\multicolumn{3}{l}{\textbf{Reverse curriculum: Low $\rightarrow$ Medium $\rightarrow$ High}} \\
CURE-MED-Qwen2.5-1.5B & 33.48 $\pm$ 0.86 & 15.92 $\pm$ 0.71 \\
CURE-MED-Qwen2.5-3B   & 47.76 $\pm$ 0.79 & 23.64 $\pm$ 0.66 \\
CURE-MED-Qwen2.5-7B   & 64.21 $\pm$ 0.73 & 36.58 $\pm$ 0.60 \\
\midrule
\multicolumn{3}{l}{\textbf{Representative size matched baselines}} \\
Qwen2.5-Instruct-1.5B & 3.84 $\pm$ 0.25 & 6.20 $\pm$ 0.24 \\
Qwen2.5-Instruct-3B   & 8.39 $\pm$ 0.42 & 10.83 $\pm$ 0.60 \\
Qwen2.5-Instruct-7B   & 25.44 $\pm$ 0.36 & 29.56 $\pm$ 0.42 \\
\bottomrule
\end{tabular}%
}
\caption{Curriculum ordering ablation. We report mean $\pm$ standard deviation over three runs for language consistency and logical accuracy.}
\label{tab:curriculum_order_half}
\end{table}

\end{document}